%% file: main.tex
\documentclass[letterpaper]{article} 
\usepackage{aaai25}  
\usepackage{times}  
\usepackage{helvet}  
\usepackage{courier}  
\usepackage[hyphens]{url}  
\usepackage{graphicx} 
\usepackage{natbib}  
\usepackage{caption} 
\usepackage{algorithm}
\usepackage{algorithmicx}
\usepackage{algpseudocode}
\usepackage{amsmath}

\usepackage{xcolor}
\newcommand{\changed}[1]{\textcolor{black}{#1}}

\usepackage{newfloat}
\usepackage{listings}

\DeclareCaptionStyle{ruled}{labelfont=normalfont,labelsep=colon,strut=off} 
\floatstyle{ruled}
\newfloat{listing}{tb}{lst}{}
\floatname{listing}{Listing}
\title{Sim911: Towards Effective and Equitable 9-1-1 Dispatcher Training with an LLM-Enabled Simulation}
\author {
    Zirong Chen\textsuperscript{\rm 1},
    Elizabeth Chason\textsuperscript{\rm 1},
    Noah Mladenovski\textsuperscript{\rm 2},
    Erin Wilson\textsuperscript{\rm 2},\\
    Kristin Mullen\textsuperscript{\rm 2},
    Stephen Martini\textsuperscript{\rm 2},
    Meiyi Ma\textsuperscript{\rm 1}
}
\affiliations {
    \textsuperscript{\rm 1}Department of Computer Science, Vanderbilt University, Nashville, Tennessee 37235, USA\\
    \textsuperscript{\rm 2}Metro Department of Emergency Communications, Nashville, Tennessee 37211, USA\\
        \{zirong.chen, elizabeth.r.chason, meiyi.ma\}@vanderbilt.edu\\
        \{noah.mladenovski, erin.wilson, kristin.mullen, stephen.martini\}@nashville.gov
}

\usepackage{multicol}
\usepackage{xcolor} 
\usepackage{multirow}
\usepackage{soul}
\usepackage{tabularx} 
\usepackage{amssymb}  
\usepackage{tabularray}
\usepackage{pifont}

\usepackage{amsmath}

\begin{document}

\sethlcolor{black}

\newcommand{\cmark}{\ding{51}}%
\newcommand{\xmark}{\ding{55}}%

\definecolor{betterblue}{rgb}{0.298, 0.608, 0.902}
\definecolor{betterred}{rgb}{0.859, 0.157, 0.204}
\definecolor{betteryellow}{rgb}{0.929, 0.702, 0.118}
\definecolor{bettergreen}{rgb}{0.012, 0.373, 0.216}

\raggedbottom

\maketitle

\begin{abstract}
Emergency response services are vital for enhancing public safety by safeguarding the environment, property, and human lives. 
As frontline members of these services, 9-1-1 dispatchers have a direct impact on response times and the overall effectiveness of emergency operations.
However, traditional dispatcher training methods, which rely on role-playing by experienced personnel, are labor-intensive, time-consuming, and often neglect the specific needs of underserved communities. 
To address these challenges, we introduce Sim911\changed{\footnote{More details: \url{https://meiyima.github.io/angie.html}}}, the first training simulation for 9-1-1 dispatchers powered by Large Language Models (LLMs). Sim911 enhances training through three key technical innovations:
(1) knowledge construction, which utilizes archived 9-1-1 call data to generate simulations that closely mirror real-world scenarios; 
(2) context-aware controlled generation, which employs dynamic prompts and vector bases to ensure that LLM behavior aligns with training objectives; and
(3) validation with looped correction, which filters out low-quality responses and refines the system performance.
Beyond its technical advancements, Sim911 delivers significant social impacts. Successfully deployed in \changed{the Metro Nashville Department of Emergency Communications (MNDEC)}, Sim911 has been integrated into multiple training sessions, saving time for dispatchers. By supporting a diverse range of incident types and caller tags, Sim911 provides more realistic and inclusive training experiences. In a conducted user study, 90.00\% of participants found Sim911 to be as effective or even superior to traditional human-led training, making it a valuable tool for emergency communications centers nationwide, particularly those facing staffing challenges.
\end{abstract}

%

\input{1_introduction}

\input{2_motivation}

\input{3_approach}
\input{4_experiments}

\input{5_related_work}
\input{6_summary}
\input{7_acknowledgement}

\bibliography{refs}


\input{_appendix}
\end{document}

%% file: 1_introduction.tex
\section{Introduction}


Emergency response services are essential for public safety, managing around 240 million 911 calls annually, based on year-round stats from New York City \cite{nycnextgen911}. However, there is a critical staffing shortfall, with a third of centers reporting more vacancies in 2023 compared to 2019, resulting in approximately 25,000 unfilled positions nationwide. This staffing crisis increases the workload on current staff \changed{\cite{chen2022cityspec, chen2023cityspec}}, leading to dispatcher burnout and impacting emergency service quality \cite{NENA2024}. As urban areas in the US grow, the strain on emergency response systems intensifies. Rapid urbanization and population growth demand effective solutions to adapt to and manage these increasing pressures \cite{ma2019data}.


Traditional training environments prepare trainees for real-world dispatcher roles by employing role-playing scenarios where experienced dispatchers coach trainees through simulated calls. The need for experienced dispatchers to participate in training diverts essential personnel from actual emergency duties, causing inconsistencies in training quality and reducing the availability of skilled staff, particularly in underserved areas \cite{2021annual, g12afonso2021planning}. However, traditional training methods, which rely on role-playing by experienced personnel, are labor-intensive, time-consuming, and frequently overlook the specific needs of underserved communities.



In light of these demands, exploring innovative technological solutions is critical. Advancements in artificial intelligence, especially Large Language Models (LLMs), offer promising methods for enhancing training environments. Employing LLMs to simulate caller interactions can reduce reliance on human resources, improving training efficiency and consistency \cite{naveed2023comprehensive, wang2023aligning, carta2023grounding}. 
However, directly applying plain LLM agents is not ideal. In our preliminary investigations, we identified the following \textbf{challenges}:
(1) Achieving consistent \textit{realistic} simulations is difficult without detailed factual databases, despite meticulous prompt engineering across different LLMs. This lack of realism results in simulations that do not fit the local context, making the training less effective and potentially confusing for trainees.
(2) LLMs excel in generating coherent content but tend to fabricate details, undermining \textit{authenticity}. Simulations with fabricated geographic information lead dispatchers to make decisions based on incorrect data, compromising emergency response effectiveness.
(3) The needs of \textit{vulnerable populations} in metropolitan areas are often understudied during conventional training, leaving practitioners unprepared. This lack of inclusiveness results in biased training, inadequately preparing dispatchers for handling calls from vulnerable groups, and leading to disparities in emergency response.
(4) The inherently complex nature of 9-1-1 calls presents significant challenges, even for human trainers. Human-led training may also fail to capture these complexities, as discussed in detail in Section \ref{sec:motivation}.

In this paper, we introduce Sim911, the first system that leverages LLMs to simulate realistic 9-1-1 calls, specifically designed to enhance dispatcher training. Sim911 focuses on creating effective and equitable simulation experiences tailored to the local metro area. Sim911 comprises three key components: \textit{knowledge construction}, \textit{context-aware controlled generation}, and \textit{validation with looped feedback}. Knowledge construction organizes real-world information into retrieval knowledge bases, while context-aware controlled generation fine-tunes the LLM's behavior through human-designed instructions. Validation with looped feedback ensures high-quality outputs by filtering out low-quality responses.

We summarize our \textbf{technical innovations} and \textbf{contributions} as follows:
(1) \textit{Innovative Knowledge Construction from 9-1-1 Calls}: Sim911 organizes real-world call data into detailed knowledge bases, allowing for the generation of contextually accurate and realistic training simulations, supporting 57 different incident types.
(2) \textit{Context-Aware Controlled Generation}: Sim911 strategically and dynamically uses advanced techniques, such as Chain-of-Thought (CoT) and Retrieval-Augmented Generation (RAG), to tailor LLM behavior during training sessions.
(3) \textit{Validation with Looped Correction}: Sim911 includes a unique validation process that filters out low-quality responses, ensuring high-quality and scenario-appropriate outputs.
(4) \textit{Focus on Social Equity}: Sim911 emphasizes training that addresses the needs of underserved and vulnerable communities, incorporating relevant data to better prepare dispatchers for diverse real-world scenarios.
(5) \textit{Real-World Deployment and Evaluation}: Successfully deployed in DEC, Sim911 has proven an effective tool in enhancing 9-1-1 dispatcher training from experimental results on real-world data.

Beyond technical advancements, Sim911 delivers significant \textbf{social impacts}:
(1) Sim911 has been successfully deployed in DEC's training programs, seamlessly integrated into 4 training classes across different service sites.
(2) To the date of this paper, Sim911's system logs reveal a total active simulation time of 26.55 hours, effectively saving this time for MNDEC dispatchers.
(3) Sim911 supports 57 different real-world incident types and covers 14 caller tags, such as ``unhoused'' and ``non-English speaking,'' to enrich caller profiles.
(4) In a user study conducted with DEC, 90.00\% of participants found Sim911 to be as effective or even superior to traditional human-led training. Additionally, Sim911 received an average helpfulness score of 4.89 for its assistance in call-taking training.
(5) Sim911 has the potential to assist emergency communications centers across the U.S. with limited staffing by enabling trainees to engage individually with the training program.

%% file: 2_motivation.tex
\section{Motivating Study}
\label{sec:motivation}



We analyzed 11,841 real-world phone call recordings (from Nov. 2022 to May 2024) and manually reviewed 33 conventional training pieces (see examples in the Appendix, leading to the following observations.


\noindent \textbf{Traditional training is laborious and time-consuming}.
In traditional training setups, each trainee engages in call simulations, assuming three roles: the call-taker, the caller, and the instructor. The trainee, as the call-taker, manages calls directed by the instructor and an experienced dispatcher. Each trainee typically participates in 60 independent simulated calls, with each call requiring the involvement of three participants. Based on past recordings, the average call duration is 3.5 minutes; with an average of 12 trainees per session, the total time commitment for experienced dispatchers amounts to at least 84 working hours per session.

\noindent \textbf{Real-world 9-1-1 calls cover a wide spectrum of incident types and contextual scenarios}. 
From our analysis of past phone call recordings, we identified over 200 distinct incident specifications. However, during initial training, each trainee is exposed to only 40 incident types and 15 call templates. Our review reveals that, on average, trainees cover only 48.00\% of the incident types in the first 3-day program, and only 61.54\% of special contexts or requests are adequately addressed. This limited exposure fails to prepare trainees for the variety of incidents they will encounter.

\begin{figure}
    \centering
    \includegraphics[width=0.9\linewidth]{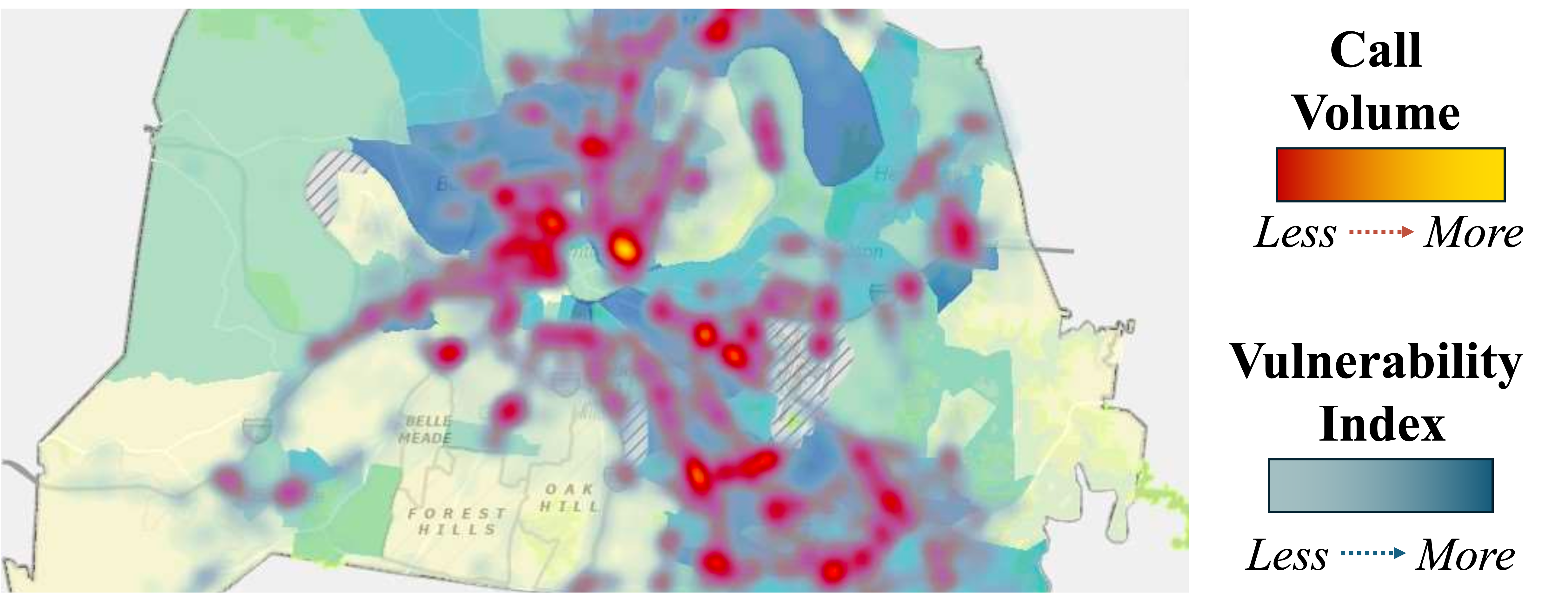}
    \caption{\small{Year-round Distribution of 9-1-1 Calls and Vulnerability Index in Nashville, 2023. 
    }}
    \label{fig:nash}
\end{figure}

\noindent \textbf{Caller images are critical for call-taking training but rarely considered}.
Equity and inclusiveness are often overlooked in conventional role-playing simulations during dispatcher training. Even with guidance from experienced dispatchers, these simulations frequently struggle to empathetically and accurately capture the nuanced experiences of vulnerable groups. Among the 33 training scenarios we reviewed, only 4 focused on vulnerable populations (such as non-native English speakers, who may use different language patterns; and callers from lower-income housing areas, who might have limited access to personal vehicles) representing just 12.12\%. However, government statistics, see Figure \ref{fig:nash}, indicate that the needs of various vulnerable groups are significantly reflected in real-world 9-1-1 calls. This discrepancy highlights the importance of incorporating diverse caller images into training, as different scenarios might require distinct call-taking skill sets to effectively handle real-life situations.



%% file: 3_approach.tex
\section{Methodology}

\begin{figure*}[t] 

\centering
\includegraphics[width=0.9\textwidth, keepaspectratio=true]{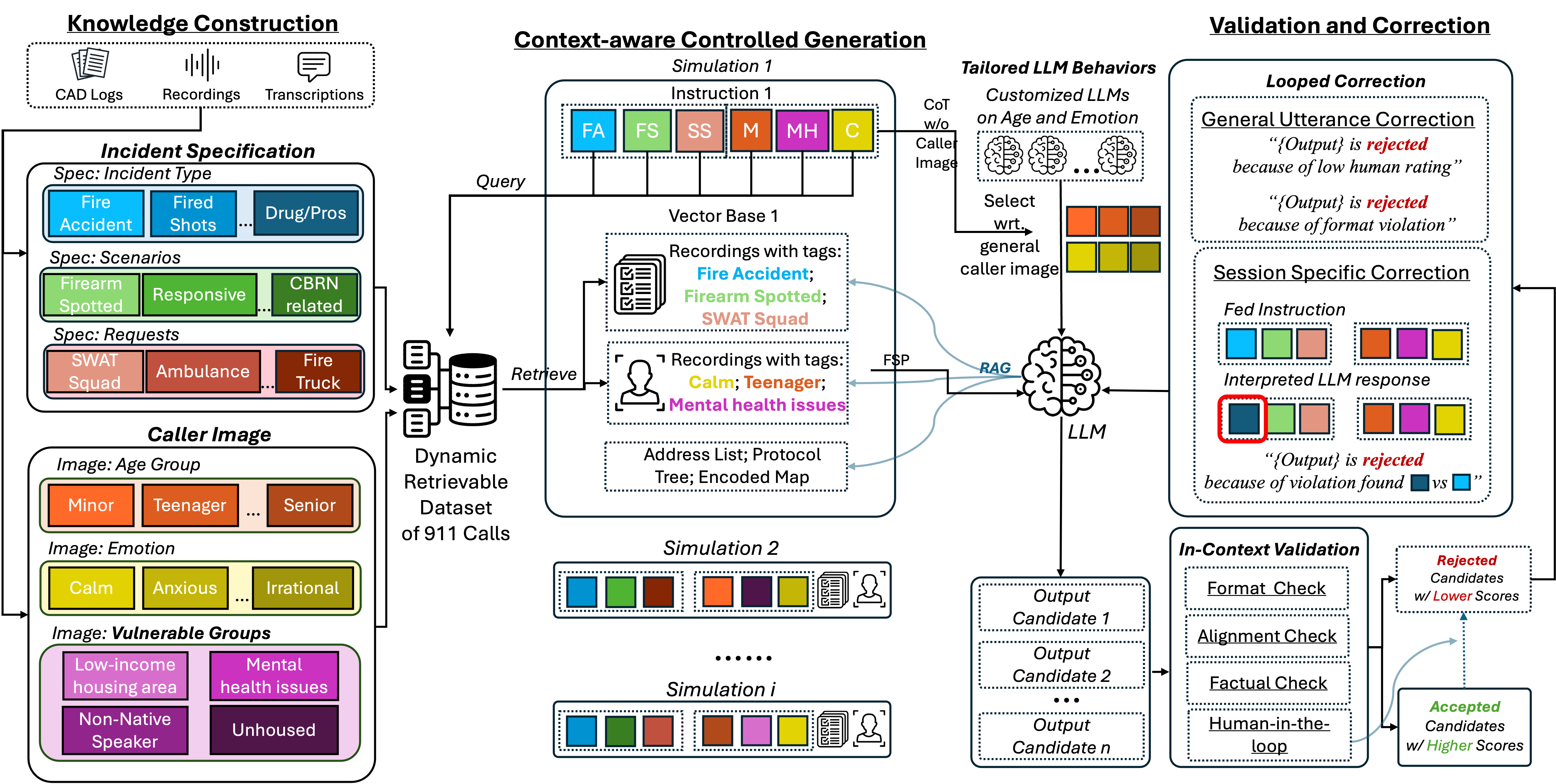}
\caption{\small{An Overview of Sim911's Key Components: \textit{knowledge construction} (KC), \textit{context-aware controlled generation} (CaCG), and \textit{validation with looped correction} (VLC). KC integrates real-world data into knowledge bases before runtime. CaCG tailors LLM behaviors during runtime. VLC involves in-context validation during runtime and periodic correction after runtime.}} \label{fig:llm}

\end{figure*}

This section first provides an overview of Sim911. Then we introduce the technical aspects of how Sim911 works in Sections Knowledge Construction, Context-aware Controlled Generation, and Validation with Looped Correction. 

Sim911 simulates calls by playing the role of 9-1-1 callers and interacting directly with the trainees. It comprises three main components, depicted in Figure \ref{fig:llm}: \textit{knowledge construction}, \textit{context-aware controlled generation}, and \textit{validation with looped correction}. During the \textit{Knowledge Construction} phase, 11,841 calls are analyzed to develop knowledge bases containing tags for incident specifics and caller images. This ensures that pertinent information is readily available during simulations. At runtime, Sim911 utilizes these tags to select the most suitable LLM backends, query the knowledge bases, and generate prompts in the \textit{context-aware controlled generation} phase. This process tailors prompts to include incident details and caller profiles, ensuring that LLM responses align with simulation requirements. The \textit{Validation with Looped Correction} component filters out low-quality responses in as the simulation goes.

\subsection{Dynamic Knowledge Construction}
\label{sec:knowledge}

\begin{figure*}[h] 

\centering
\includegraphics[width=\textwidth, keepaspectratio=true]{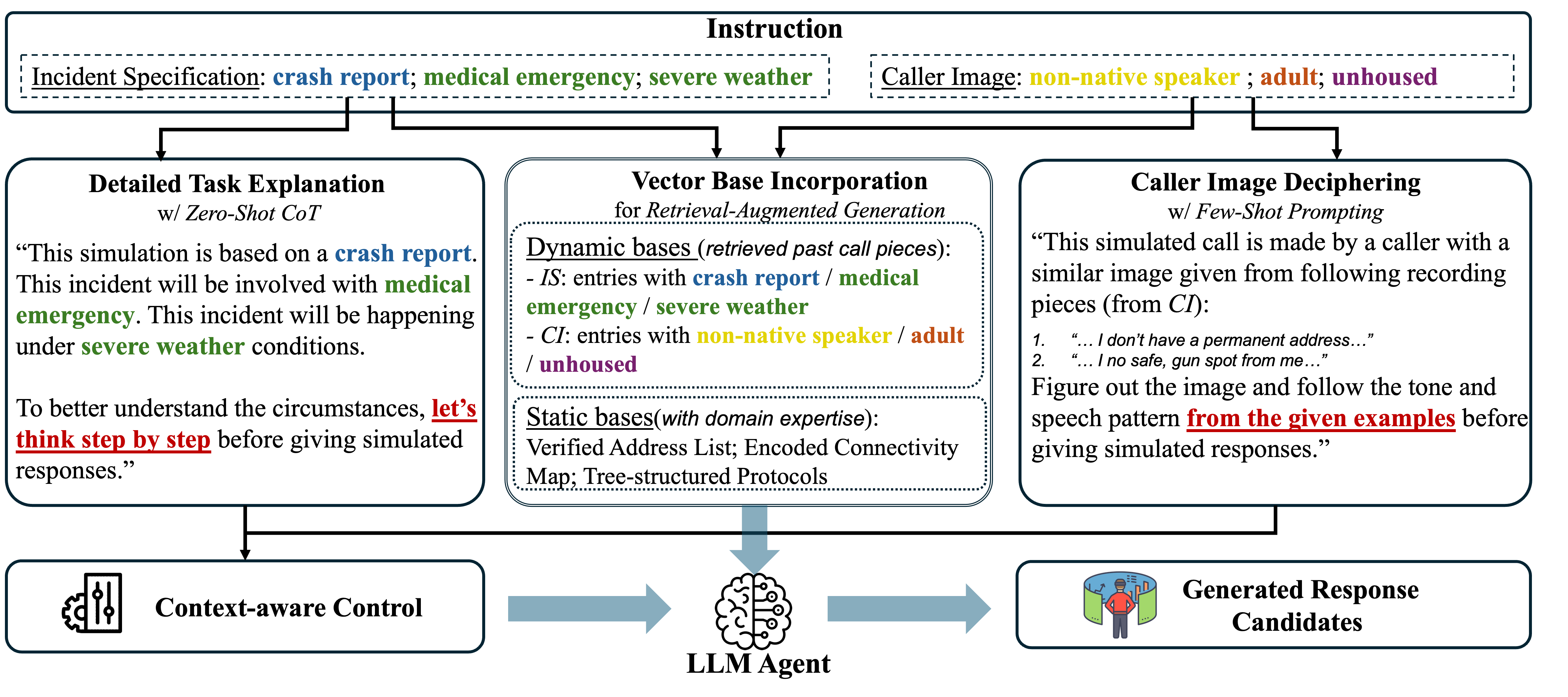}
\caption{\small{\changed{An Example of the 3-step \textit{Context-aware Controlled Generation} with Vector Base Incorporation, Detailed Task Explanation, and Caller Image Deciphering}. The incident type specification (\textit{IS}) comes with tags \textcolor{betterblue}{\textbf{crash report}} (incident type), \textcolor{bettergreen}{\textbf{medical emergency}}(special request), and \textcolor{bettergreen}{\textbf{severe weather}} (scenario contexts), and the caller image (\textit{CI}) is set to be an \textcolor{purple}{\textbf{unhoused}} \textcolor{betteryellow}{\textbf{non-native speaker}} \textcolor{betterred}{\textbf{adult}}.}} \label{fig:prompt}


\end{figure*}

As a first step, we build an in-depth review and sophisticated reconstruction of the existing dataset, see running examples in Appendix, which has three key data sources: Computer-Aided Dispatching(CAD) logs, archived 9-1-1 call recordings, and their corresponding transcriptions. 



\subsubsection{Detailing Two Components in 9-1-1 Calls}

We integrate insights from dispatcher teams at MNDEC to identify two key components in 9-1-1 call handling: \textit{Incident Specifications} and \textit{Caller Images}. We use finely-grained tags for each data entry to create more accurate simulations. In our annotation work, we manually review each call and apply all relevant tags. A single call may be annotated with multiple tags to ensure comprehensive coverage.

\textit{Incident Specifications} (\textit{IS}) capture critical details of incidents, including:
(1) \textit{Incident Type}: Categorizes the incident, from routine (e.g., illegal parking) to critical (e.g., severe medical emergencies).
(2) \textit{Scenario Context}: Adds situational context, such as environmental conditions (e.g., severe weather), potential threats (e.g., sightings of firearms), or specific events (e.g., large public gatherings).
(3) \textit{Special Requests}: Identifies specific instructions, like the need for specialized units (e.g., bomb squads) or coordination with other agencies (e.g., fire departments).


\textit{Caller Images} (\textit{CI}) create a comprehensive caller profile, enhancing the LLM's understanding of the caller's perspective, especially for vulnerable groups. This includes:
(1) \textit{General Tags}: Profiles the caller by age (e.g., minor, adult) and emotion (e.g., ``neutral'', ``anxious"). Dispatchers assign these tags based on conversation clues (e.g., ``My mom is mid-70s and living alone") or voice analysis. These tags are less sensitive and linked to pre-customized LLM agents to avoid identification confusion \cite{wei2024jailbroken}.
(2) \textit{Vulnerable Groups}: These government-introduced tags include descriptors such as ``low-income housing area'' (if the call originates from a lower-income area, according to year-round statistics), ``mental health'' (if the caller exhibits potential mental health issues, such as bipolar disorder or depression; inferred from the conversation), ``non-native speaker'' (if the caller uses limited English), and ``unhoused'' (if the caller indicates lack of stable, permanent housing, inferred from conversation clues). These tags are considered highly sensitive and remain hidden during runtime due to ethical concerns.

\subsubsection{Specializing Knowledge Bases for Contextual Control}

We leverage the Retrieval Augmented Generation (RAG) approach, which enhances LLMs for tasks requiring deep knowledge by incorporating external databases as reference points during content generation. This methodology, as discussed by \cite{lewis2020retrieval}, improves the LLMs' ability to provide accurate and relevant outputs. Here, we introduce the two major knowledge bases for runtime use.

\textit{Factual Bases}. First, we build a static base, which contains factual knowledge: (1) \textit{Validated Address List}: a comprehensive list of real addresses within the local area; (2) \textit{Encoded Map with Connectivity Information}: beyond simple address listings, this map provides detailed information about the connectivity between locations; (3) \textit{Tree-Structured Protocols}: a collection of protocols for various types of emergency incidents, organized in a tree structure. These protocols detail the question sequence dispatchers should follow, ensuring Sim911’s simulations adhere to procedural standards of emergency response.

\textit{Retrievable Bases}. This base includes data entries tagged according to Incident Specifications (\textit{IS}) and Caller Images (\textit{CI}). The retrievable base allows Sim911 to query and retrieve necessary data samples that enhance the simulation experience during runtime.






\subsection{Context-aware Controlled Generation}
\label{sec:control}

Each simulation runtime begins with predefined tags (referred to as `instructions') that detail the desired scenarios. These tags delineate Incident Specifications (\textit{IS}) and Caller Images (\textit{CI}), guiding the setup for each simulation. We select the most appropriate preset backend for each simulation based on less sensitive \textit{CI} attributes (emotion, age) and gather data associated with these tags from knowledge bases. This ensures a well-informed simulation environment tailored to the given simulation instructions. Context-aware Controlled Generation employs advanced prompting techniques, including Chain-of-Thought (CoT), Retrieval-Augmented Generation (RAG), and Few-shot Prompting (FSP), to enhance LLM performance \cite{touvron2023llama, kaplan2020scaling, wei2022chain}. Unlike the direct and static application of these techniques, our approach dynamically adapts to the emergency response context, with a focus on both \textit{IS} and \textit{CI}. The context-aware controlled generation process consists of three major steps, illustrated in the running example in Figure \ref{fig:prompt}:
(1) \textit{Vector Base Incorporation}: We statically mount the fact bases and dynamically retrieve all past call pieces associated with given tags in both \textit{IS} (e.g., crash report, medical emergency, severe weather) and \textit{CI} (e.g., non-English speaking, adult, unhoused). The LLM backend is granted access to both vector bases;
(2) \textit{Detailed Task Explanation}: We elaborate on the instruction through step-by-step explanations, setting the stage for how the simulation should proceed. This preparation allows the LLM to conceptualize the simulation’s context and objectives before initiation;
(3) \textit{Caller Image Deciphering}: By including examples of past utterances and interactions associated with both similar tags, we provide the LLM with contextually relevant examples to draw from. This repository of past interactions enriches the LLM’s understanding and ability to generate responses that are both consistent with the user’s profile and grounded in real-world examples.
See detailed algorithmic description in Appendix.

\subsection{Validating with Looped Correction}

During runtime, Sim911 employs an in-context validation process with a co-pilot design \changed{\cite{chen2024auto311}} to prevent negative examples and iteratively loops back to the LLM backend until a validated response is obtained. To avoid infinite loops and reduce latency, \changed{threshold of 3 is applied considering both latency and accuracy under deployment, see detailed discussion in Appendix.} The \textit{In-Context Validation} process includes four key checks:
(1) \textit{Format check}: This check ensures that the generated response adheres to the expected format. Any response that violates the format requirements is discarded to prevent system errors.
(2) \textit{Alignment check}: This step utilizes a BERT-based classifier \cite{devlin2019bert} to extract incident specifications from the response. The extracted specifications are then compared with the expected instructions, and any misalignment results in the response being discarded.
(3) \textit{Factual check}: A RoBERTa-based question-answering framework \cite{liu2019roberta} is used to query key details, such as location information, by asking preset questions (e.g., ``What is the address?''). If the extracted address does not exist in real life, the response is discarded.
(4) \textit{Human-in-the-Loop check}: This step allows users to provide immediate feedback on the generated response, supporting both written comments and scaled (1-5) ratings. Users can reject any response that does not meet their standards, and this feedback is systematically collected for further analysis.
See details in Appendix.

%% file: 4_experiments.tex
\section{Evaluation of Sim911}
Sim911 introduces a pioneering AI-driven system to enhance call-taker training for emergency response scenarios. Due to its novelty, there is limited existing literature to guide its evaluation. To provide a comprehensive assessment, we not only report system-level performance but also conduct a study on the component-wise impacts using pre-configured runtimes. This approach allows us to evaluate Sim911 component by component without disrupting its ongoing deployment at DEC. Therefore, our evaluation of Sim911’s performance, focusing on \textbf{\textit{effectiveness}} and \textbf{\textit{equity}}, consists of two components: (1) component-wise analysis using pre-configured runtimes, and (2) system-level assessment during real-world deployment. \changed{GPT-4o is tested to be the optimal LLM backend for Sim911 by the date of submission. Refer to the complete baseline comparison in Appendix.} We fetched GPT-4o responses using OpenAI API and tested the workflow on a machine with a 2.50GHz CPU, 32GB RAM, and Nvidia RTX 3080Ti GPU.

\textit{Component-wise analysis with pre-configured runtimes}: We begin by extracting instructions from 2,641 past calls in the MNDEC database, spanning 13 incident types (e.g., Motor Vehicle Accidents 11\%, Lost and Stolen 10\%, Aggressive Drivers 10\%) from Nov. 2022 to May 2024, based on Incident Specifications (\textit{IS}) and Caller Images (\textit{CI}) provided by expert annotations. We then replicate the dialogue flows using rule-based scripts that mimic the questions typically asked by call-takers. These instructions and replicated dialogue flows are used to simulate calls with Sim911. Sim911 operates without access to highly granular details. For example, if a past call involved an abandoned vehicle with a specific license plate and tinted windows, Sim911 would only be informed that the simulation involves an abandoned vehicle, without further specifics like the license plate or tinted windows. To ensure a fair evaluation, we exclude these granular discrepancies from our analysis. Effectiveness and equity scores are obtained through call-wise comparisons under control experiments. We record average scores with standard deviations to ensure robust evaluation.

\begin{table*}[ht]
\renewcommand{\arraystretch}{1.1}
\centering
\small
\resizebox{0.75\textwidth}{!}{%
\begin{tabular}{|cc|ccc|cc|}
\hline
\multicolumn{2}{|c|}{\multirow{2}{*}{}}                            & \multicolumn{3}{c|}{\textbf{\textsc{Realism}}}                                                              & \multicolumn{2}{c|}{\textbf{\textsc{Authenticity}}} \\ \cline{3-7} 
\multicolumn{2}{|c|}{}                                             & \multicolumn{1}{c|}{PPL$\downarrow$} & \multicolumn{1}{c|}{METEOR$\uparrow$} & TTR$\uparrow$   & \multicolumn{1}{c|}{GMap(\%)$\uparrow$} & SAR(\%)$\uparrow$ \\ \hline\hline
\multicolumn{2}{|c|}{Sim911$\neg$KC}                               & \multicolumn{1}{c|}{31.22$\pm$13.38} & \multicolumn{1}{c|}{0.22$\pm$0.05}    & 0.88$\pm$0.01  & \multicolumn{1}{c|}{83.00$\pm$5.11}     & 94.11$\pm$1.66    \\ \hline
\multicolumn{1}{|c|}{\multirow{3}{*}{Sim911$\neg$CaCG}} & $\neg$CoT & \multicolumn{1}{c|}{21.98$\pm$6.64}  & \multicolumn{1}{c|}{0.67$\pm$0.18}    & 0.85$\pm$0.02  & \multicolumn{1}{c|}{98.13$\pm$1.06}     & 90.83$\pm$2.71    \\ \cline{2-7} 
\multicolumn{1}{|c|}{}                                 & $\neg$FSP & \multicolumn{1}{c|}{12.55$\pm$5.71}  & \multicolumn{1}{c|}{0.75$\pm$0.21}    & 0.94$\pm$0.01  & \multicolumn{1}{c|}{98.11$\pm$1.08}     & 98.14$\pm$1.06    \\ \cline{2-7} 
\multicolumn{1}{|c|}{}                                 & $\neg$RAG & \multicolumn{1}{c|}{57.19$\pm$12.22} & \multicolumn{1}{c|}{0.19$\pm$0.12}    & 0.92$\pm$0.02  & \multicolumn{1}{c|}{61.47$\pm$6.89}     & 96.44$\pm$2.60    \\ \hline
\multicolumn{2}{|c|}{Sim911$\neg$VLC}                              & \multicolumn{1}{c|}{18.89$\pm$8.19}  & \multicolumn{1}{c|}{0.77$\pm$0.08}    & 0.88$\pm$0.02  & \multicolumn{1}{c|}{89.11$\pm$5.15}     & 89.48$\pm$3.12    \\ \hline
\multicolumn{2}{|c|}{Sim911$\neg$All(\textit{GPT-4o})}                              & \multicolumn{1}{c|}{61.99$\pm$13.91} & \multicolumn{1}{c|}{0.12$\pm$0.02}    & 0.88$\pm$0.02  & \multicolumn{1}{c|}{61.01$\pm$11.31}    & 81.63$\pm$3.97    \\ \hline
\multicolumn{2}{|c|}{\textbf{Sim911}}                              & \multicolumn{1}{c|}{\textbf{11.07$\pm$5.49}}  & \multicolumn{1}{c|}{\textbf{0.85$\pm$0.03}}    & \textbf{0.94$\pm$0.01} & \multicolumn{1}{c|}{\textbf{99.19$\pm$0.81}}     & \textbf{98.42$\pm$1.58}    \\ \hline
\end{tabular}%
}
\caption{\small{\textbf{\textit{Effectiveness}} of Sim911 in terms of \textsc{\textbf{Realism}} and \textsc{\textbf{Authenticity}}. The metrics used include \textbf{\textit{PPL}} (Perplexity), \textbf{\textit{TTR}} (Type-Token Ratio), \textbf{\textit{GMap}} (Google Maps API locating success rate), and \textbf{\textit{SAR}} (Simulation Alignment Rate from copilot's results).}}
\label{tab:effective}
\end{table*}

\textit{System-level assessment during real-world deployment}: During Sim911's deployment, we collected data from 3,416 system interactions and 3,409 user interactions across both complete (228) and incomplete simulations, each guided by its own set of instructions (\textit{IS} and \textit{CI}). These data are utilized in assessing Sim911's authenticity. Additionally, we conducted a user study in collaboration with MNDEC to evaluate Sim911 at a system level. This study involved trainees and personnel from DEC, including those from training management and quality assurance. The user study gathered scaled feedback (1-5) from MNDEC personnel on several key aspects, e.g., \textit{realism} (`How similar or vivid are the calls generated by Sim911 compared to real-world calls?'), \textit{authenticity} (`Are Sim911’s responses valid and true to real-life situations?'), \textit{equity} (`How well does Sim911 simulate the experiences of vulnerable callers?'), and \textit{helpfulness} (`How helpful is Sim911 in assisting with call-taking training?'). Written comments were also collected to provide additional insights. See survey details in Appendix. To assess effectiveness and equity, we review system logs and user feedback, with further details discussed later.

\subsection{Effectiveness of Sim911}
We assess Sim911's effectiveness by following aspects: \textbf{\textit{realism}} and \textbf{\textit{authenticity}}.

\noindent \textit{\textbf{Realism}}: \textit{``How closely do Sim911’s simulations mirror real-world calls?''} We use the following metrics to evaluate Sim911’s performance on pre-configured runtimes: \textit{Perplexity} (a measure of distributional similarity commonly used in language model training; it assesses how reasonable the generated texts are compared to a reference set), \textit{METEOR} \cite{banerjee2005meteor} (text generation metric that balances precision and recall, considering word stems, synonyms, and word order to determine how closely a generated text mirrors a reference text), and \textit{TTR} (Type-Token Ratio; measures lexical diversity by comparing the number of unique words to the total number of words in the text).

\noindent \textit{\textbf{Authenticity}}: \textit{``Does Sim911 provide accurate, true-to-life information without fabricating given instructions?''} For evaluation, we break authenticity down into two aspects: ``matter of facts'' and ``simulation alignment.'' For the first, we focus on the accuracy of the given location in a simulation, as recommended by MNDEC experts. We use the Google Maps API with Geocoding \cite{GoogleMapsPlatform} to verify the geographic information provided in the simulation and report the \textit{locating success rate}. To measure \textit{simulation alignment}, we use the copilot’s results to determine if the indicated Incident Specification (\textit{IS}) aligns with the one provided in the simulation instructions. System-level performance during real-world deployment is assessed through quantitative analysis of system logs.

From the statistics in Table \ref{tab:effective}, we observe the following key points. When all components are enabled, Sim911 achieves optimal results in both realism (PPL=11.07, METEOR=0.85) and authenticity (GMap=99.19\%, SAR=98.42\%). Disabling knowledge construction (KC) and the RAG sub-component of CaCG leads to significant drops in realism (PPL=31.22 and PPL=57.19, respectively). Similarly, turning off validation with looped correction (VLC) reduces both realism and authenticity, though the system remains moderately effective. When all components are disabled, the system's performance declines significantly, particularly in realism (PPL=61.99) and authenticity (SAR=81.63\%). In conclusion, \textit{Sim911 demonstrates high effectiveness in terms of realism and authenticity in real-world deployment when all components are active. Disabling components harms Sim911's overall effectiveness.}


\begin{table*}[ht]
\centering
\small
\renewcommand{\arraystretch}{1.2}
\resizebox{0.88\textwidth}{!}{%
\begin{tabular}{|cc|cc|ccc|cc|}
\hline
\multicolumn{2}{|c|}{\multirow{2}{*}{}}                             & \multicolumn{2}{c|}{\textbf{\textsc{Age Groups}}}        & \multicolumn{3}{c|}{\textbf{\textsc{Emotion Ranges}}}                                                   & \multicolumn{2}{c|}{\textbf{\textsc{Unhoused}}}           \\ \cline{3-9} 
\multicolumn{2}{|c|}{}                                              & \multicolumn{1}{c|}{BART(\%)$\uparrow$}      & Margin$\uparrow$       & \multicolumn{1}{c|}{BART(\%)$\uparrow$}      & \multicolumn{1}{c|}{Margin$\uparrow$}       & NRCLex$\uparrow$        & \multicolumn{1}{c|}{BART(\%)$\uparrow$}       & Margin$\uparrow$       \\ \hline \hline
\multicolumn{2}{|c|}{Sim911$\neg$KC}                                & \multicolumn{1}{c|}{66.46$\pm$3.89}          & 0.13$\pm$0.11          & \multicolumn{1}{c|}{71.68$\pm$3.55}          & \multicolumn{1}{c|}{0.19$\pm$0.14}          & 69.85$\pm$2.61          & \multicolumn{1}{c|}{59.87$\pm$4.45}           & 0.11$\pm$0.21          \\ \hline
\multicolumn{1}{|c|}{\multirow{3}{*}{Sim911$\neg$CaCG}} & $\neg$CoT & \multicolumn{1}{c|}{78.65$\pm$2.95}          & 0.29$\pm$0.15          & \multicolumn{1}{c|}{86.12$\pm$3.56}          & \multicolumn{1}{c|}{0.34$\pm$0.23}          & \textbf{77.11$\pm$3.44} & \multicolumn{1}{c|}{72.44$\pm$3.19}           & 0.20$\pm$0.10          \\ \cline{2-9} 
\multicolumn{1}{|c|}{}                                  & $\neg$FSP & \multicolumn{1}{c|}{62.90$\pm$3.51}          & 0.09$\pm$0.06          & \multicolumn{1}{c|}{63.41$\pm$4.71}          & \multicolumn{1}{c|}{0.08$\pm$0.06}          & 65.97$\pm$3.12          & \multicolumn{1}{c|}{60.11$\pm$3.51}           & 0.01$\pm$0.20          \\ \cline{2-9} 
\multicolumn{1}{|c|}{}                                  & $\neg$RAG & \multicolumn{1}{c|}{64.63$\pm$4.13}          & 0.13$\pm$0.12          & \multicolumn{1}{c|}{68.11$\pm$3.78}          & \multicolumn{1}{c|}{0.22$\pm$0.19}          & 61.00$\pm$3.10          & \multicolumn{1}{c|}{61.84$\pm$5.76}           & 0.05$\pm$0.16          \\ \hline
\multicolumn{2}{|c|}{Sim911$\neg$VLC}                               & \multicolumn{1}{c|}{78.71$\pm$4.29}          & 0.22$\pm$0.17          & \multicolumn{1}{c|}{76.12$\pm$4.69}          & \multicolumn{1}{c|}{0.25$\pm$0.18}          & 64.11$\pm$3.90          & \multicolumn{1}{c|}{69.41$\pm$4.17}           & 0.15$\pm$0.14          \\ \hline
\multicolumn{2}{|c|}{Sim911$\neg$All(\textit{GPT-4o})}                               & \multicolumn{1}{c|}{59.11$\pm$5.55}          & 0.07$\pm$0.05          & \multicolumn{1}{c|}{57.87$\pm$5.86}          & \multicolumn{1}{c|}{0.21$\pm$0.19}          & 51.19$\pm$4.51          & \multicolumn{1}{c|}{51.12$\pm$5.46}           & 0.17$\pm$0.22          \\ \hline
\multicolumn{2}{|c|}{\textbf{Sim911}}                               & \multicolumn{1}{c|}{\textbf{83.11$\pm$2.82}} & \textbf{0.34$\pm$0.26} & \multicolumn{1}{c|}{\textbf{85.66$\pm$3.17}} & \multicolumn{1}{c|}{\textbf{0.36$\pm$0.22}} & 73.31$\pm$2.16          & \multicolumn{1}{c|}{\textbf{73.94$\pm$4.31}}  & \textbf{0.21$\pm$0.13} \\ \hline \hline
\multicolumn{2}{|c|}{\multirow{2}{*}{}}                             & \multicolumn{2}{c|}{\textbf{\textsc{Mental Health}}}     & \multicolumn{3}{c|}{\textbf{\textsc{Non-Native Speakers}}}                                              & \multicolumn{2}{c|}{\textbf{\textsc{Low-income Housing}}} \\ \cline{3-9} 
\multicolumn{2}{|c|}{}                                              & \multicolumn{1}{c|}{BART(\%)$\uparrow$}      & Margin$\uparrow$       & \multicolumn{1}{c|}{BART(\%)$\uparrow$}      & \multicolumn{1}{c|}{Margin$\uparrow$}       & Gunning Fog$\uparrow$   & \multicolumn{1}{c|}{BART(\%)$\uparrow$}       & Margin$\uparrow$       \\ \hline \hline
\multicolumn{2}{|c|}{Sim911$\neg$KC}                                & \multicolumn{1}{c|}{67.67$\pm$4.41}          & 0.17$\pm$0.11          & \multicolumn{1}{c|}{70.19$\pm$4.56}          & \multicolumn{1}{c|}{0.22$\pm$0.21}          & 0.02$\pm$0.17           & \multicolumn{1}{c|}{51.13$\pm$5.17}           & 0.04$\pm$0.11          \\ \hline
\multicolumn{1}{|c|}{\multirow{3}{*}{Sim911$\neg$CaCG}} & $\neg$CoT & \multicolumn{1}{c|}{81.67$\pm$5.40}          & 0.31$\pm$0.11          & \multicolumn{1}{c|}{82.13$\pm$4.87}          & \multicolumn{1}{c|}{0.45$\pm$0.10}          & 0.17$\pm$0.11           & \multicolumn{1}{c|}{74.49$\pm$4.11}           & 0.13$\pm$0.18          \\ \cline{2-9} 
\multicolumn{1}{|c|}{}                                  & $\neg$FSP & \multicolumn{1}{c|}{64.55$\pm$3.42}          & 0.11$\pm$0.14          & \multicolumn{1}{c|}{72.13$\pm$4.41}          & \multicolumn{1}{c|}{0.39$\pm$0.11}          & 0.09$\pm$0.14           & \multicolumn{1}{c|}{57.78$\pm$4.14}           & 0.07$\pm$0.11          \\ \cline{2-9} 
\multicolumn{1}{|c|}{}                                  & $\neg$RAG & \multicolumn{1}{c|}{76.71$\pm$4.45}          & 0.01$\pm$0.09          & \multicolumn{1}{c|}{75.33$\pm$5.19}          & \multicolumn{1}{c|}{0.23$\pm$0.18}          & 0.11$\pm$0.17           & \multicolumn{1}{c|}{64.65$\pm$5.77}           & 0.01$\pm$0.14          \\ \hline
\multicolumn{2}{|c|}{Sim911$\neg$VLC}                               & \multicolumn{1}{c|}{83.22$\pm$3.77}          & 0.21$\pm$0.20          & \multicolumn{1}{c|}{81.19$\pm$3.11}          & \multicolumn{1}{c|}{0.24$\pm$0.26}          & 0.07$\pm$0.09           & \multicolumn{1}{c|}{63.18$\pm$5.13}           & 0.16$\pm$0.13          \\ \hline
\multicolumn{2}{|c|}{Sim911$\neg$All(\textit{GPT-4o})}                               & \multicolumn{1}{c|}{61.62$\pm$4.15}          & 0.04$\pm$0.11          & \multicolumn{1}{c|}{69.13$\pm$4.36}          & \multicolumn{1}{c|}{0.21$\pm$0.22}          & 0.03$\pm$0.12           & \multicolumn{1}{c|}{44.44$\pm$3.72}           & 0.04$\pm$0.02          \\ \hline
\multicolumn{2}{|c|}{\textbf{Sim911}}                               & \multicolumn{1}{c|}{\textbf{86.16$\pm$3.37}} & \textbf{0.33$\pm$0.14} & \multicolumn{1}{c|}{\textbf{84.41$\pm$4.80}} & \multicolumn{1}{c|}{\textbf{0.48$\pm$0.11}} & \textbf{0.17$\pm$0.13}  & \multicolumn{1}{c|}{\textbf{77.98$\pm$3.37}}  & \textbf{0.19$\pm$0.21} \\ \hline
\end{tabular}
}
\caption{\small{Caller Image Tag-wise \textbf{\textit{Equity}} Features Analysis of Sim911. The metrics used include \textbf{\textit{BART}} (BART Score), \textbf{\textit{Margin}} (Margin Score), \textbf{\textit{NRCLex}} (accuracy on NRCLex results across tags), and \textbf{\textit{Gunning Fog}} (margin score on Gunning Fog Index across tags).}}
\label{table:equity}
\end{table*}

\subsection{Equity of Sim911}
We assess Sim911's equity features by evaluating \textit{``how effectively it provides simulation experiences for different caller groups''}, represented by each supported tag in the caller image (\textit{CI}). Recognizing that some tags are subjective and challenging to quantify, we adopt two general approaches to study these equity features. We employ fine-tuned BART \cite{lewis2019bart}, a state-of-the-art model for zero-shot text classification, to evaluate Sim911-generated emergency call texts against a predefined set of image tags. For each generated call \( x_i \), associated with ground truth tags \( \mathcal{T}(x_i) \), BART predicts the presence or absence of each tag \( T_j \) using a binary classifier \( C_j(x_i) \), which outputs 1 if \( x_i \) is associated with \( T_j \), and 0 otherwise. The predicted tags form a binary vector \( \hat{\mathcal{T}}(x_i)=\{C_1(x_i), C_2(x_i), \dots, C_k(x_i)\} \). Accuracy for each call is calculated by comparing \( \hat{\mathcal{T}}(x_i) \) with \( \mathcal{T}(x_i) \) using the formula \( \text{Acc}(x_i)=\frac{1}{k} \sum_{j=1}^{k} \mathbb{I}\left( C_j(x_i)=\mathbb{I}(t_{ij} \in \mathcal{T}(x_i)) \right) \). This classification is iteratively performed for each tag, and the overall accuracy is determined by averaging the individual accuracies across all generated calls as \( \text{\textit{BART Score}}=\frac{1}{n} \sum_{i=1}^{n} \text{Acc}(x_i) \). Second, we perform a textual similarity analysis based on syntax (Context-Free Grammar Parser), lexicon (TF-iDF), and sentiment \cite{loria2018textblob}. We compare the generated outputs tagged as \( A \) with both the ground truth tagged as \( A \) and not-\( A \). To quantify the strength of classification for each tag, we calculate the \textit{Margin Score} \( (\text{Similarity}(A) - \text{Similarity}(\neg A)) / (\text{Similarity}(A) + \text{Similarity}(\neg A)) \), where \(\text{Similarity}(.)\) is the overall syntactic similarity of the output to reference texts with a given tag.

Besides these two approaches, we additionally introduce the following tag-specific evaluation methods: (1) \textit{NRCLex} \cite{mohammad2013crowdsourcing} for unsupervised textual \textbf{\textit{emotion}} detection, where we analyze the accuracy similarly to the \textit{BART Score}. (2) \textit{Gunning Fog Index}, a well-known method in linguistics of text readability analysis, is used to assess the readability of the text for \textit{\textbf{non-native English speakers}}. Gunning Fog Index outputs a readability level and we analyze this score similarly to the \textit{Margin Score}.

From the statistics in Table \ref{table:equity}, we derive the following findings. Sim911 achieves strong performance across all caller image tags when all components are enabled, including age groups (BART=83.11\%, Margin=0.34), emotion ranges (BART=85.66\%, Margin=0.36), and unhoused populations (BART=73.94\%, Margin=0.21). Disabling the FSP sub-component of CaCG results in notable declines for age groups (BART=62.90\%, Margin=0.09) and mental health tags (BART=64.55\%, Margin=0.11). Turning off knowledge construction (KC) significantly reduces performance, especially for low-income housing (BART=51.13\%, Margin=0.04) and mental health (BART=67.67\%, Margin=0.17). Similarly, disabling validation with looped correction (VLC) leads to a decrease in metrics for mental health (BART=83.22\%, Margin=0.21). When all components are disabled, the system's performance deteriorates significantly, particularly for the low-income housing tags (BART=44.44\%, Margin=0.04) and non-native speakers (BART=69.13\%, Margin=0.21). \textit{In conclusion, Sim911 delivers equitable and inclusive simulations in real-world deployment when all components are enabled. Disabling components negatively impacts Sim911's equity features.}

\subsection{Insights from User Study}
\changed{We collected 10 anonymous feedback from trainees (x2), active call-takers/dispatchers (x2), and training officers (x6) at DEC. Surveys are contributed by MNDEC based on the availability.} Responses included yes/no questions, written comments, and a scaled rating system: Not at all (1), Neutral (2), Somewhat (3), Very much (4), and Perfectly (5). We find following insights. See complete survey setup in Appendix.

\noindent \textbf{\textit{Effectiveness and Equity}}: Sim911 received scores of 4.50 for realism and 4.70 for authenticity. In terms of equity, it performed well across various caller image tags, with average scores as follows: Age Groups (4.25), Emotion Ranges (4.20), Unhoused (4.10), Mental Health (4.25), Non-Native Speakers (4.25), and Low-Income Housing (4.10). Additionally, Sim911 earned an average score of 4.89 for \textit{``How effectively does Sim911 support call-taker training in real-life scenarios?''}. One participant commented: \textit{``I was surprised by how well it handled a call as a pregnant woman. I even managed to successfully deliver a baby on the phone!''} Another shared: \textit{``When it played the role of a kid caller, it acted just like a real child—refusing to do anything until his mom arrived on the scene.''} These results emphasize Sim911’s effectiveness in preparing call-takers by simulating diverse caller profiles and challenging real-life situations.

\noindent \textbf{\textit{Comparison to Human-led Training}}: 9 out of 10 participants found Sim911 to be on par with or better than traditional human-led training. One participant remarked: \textit{``Sim911 is a great starting point because it comes up more incident types than what we do right now. It’s a valuable tool for enhancing our training.''} Another said: \textit{``It's impressive how Sim911 can simulate different callers (images). Trainees can be exposed to rare but useful calls that we could not (simulate) in the past.''} These findings highlight that Sim911 not only complements human-led training but also enhances it by providing a broader range of incident types and scenarios that are difficult to replicate manually.

%% file: 5_related_work.tex
\section{Related Work}


\textbf{Simulation-based training} is a key component in various fields such as healthcare, aviation, and emergency services, where it provides a controlled environment for skill development without real-world risks \cite{suresh2023role, preiksaitis2023opportunities, daun2023systematic}. This method enhances critical thinking, decision-making, and practical skills by allowing repeated exposure to diverse and sometimes hazardous scenarios \cite{ibrahim2023equity, flores2023artificial, rahman2023emsreact}. Recent technological advancements, including Augmented Reality (AR), have begun to enhance traditional training setups, offering more \textbf{immersive training} experiences \cite{fitria2023augmented, pfaff2020learning, li2018learning, ummenhofer2019lagrangian}, \changed{especially those for emergency responses \cite{parry2022leadership, mehta2022human}}. Despite these innovations, most training simulations still rely heavily on human-scripted scenarios and instructor feedback, which can limit scalability and adaptability \cite{violato2023use, salvato2021crossing, de2020simulation}. \textbf{Large Language Models} (LLMs) are emerging as a transformative tool for dialogue-focused simulations, able to generate dynamic and realistic interactions \cite{webb2023proof, thoppilan2022lamda, gong2023multimodal}. But their integration into training programs must carefully address accuracy, ethical concerns, and potential biases to ensure effectiveness \cite{shanahan2023role, shayegani2023survey, yao2024survey, salewski2024context}. Refer to our extended relation work in Appendix.

%% file: 6_summary.tex
\section{Summary}

In this paper, we introduce Sim911, the first AI-driven simulation environment designed to assist 9-1-1 dispatcher training under emergency response scenarios. Sim911 aims to enhance the preparedness of emergency dispatchers, contributing to the resilience and safety of urban populations. Evaluation results on pre-configured runtimes and real-world deployment show that Sim911 effectively delivers realistic, authentic, and equitable simulations, to assist dispatcher training with the integration of knowledge construction, context-aware controlled generation, and validation with looped correction. 


This work can help emergency communications centers with limited staffing by allowing trainees to interact individually with the training program. Nearly 6,000 emergency communications centers could benefit from this training opportunity. The GenAI-enabled solution can be extended to other training spaces, such as teachers and medical students. 

%% file: 7_acknowledgement.tex
\section*{Acknowledgment}
This work was supported in part by the U.S. National Science Foundation under Grants 2427711, the Google Academic Research Award, OpenAI Researcher Access Program, and the U.S. Department of Education under Grant R305C240010.    
The opinions, findings, conclusions, or recommendations expressed in this material are those of the author(s) and do not necessarily reflect the views of the sponsoring agencies.




%% file: _appendix.tex


\clearpage


\section*{Appendix}
\appendix
\label{sec:appdendix}
\section{Discussion and Future Work}
We plan to enhance the system in several aspects. (1) \textit{Knowledge Base Timeliness}: Real-time updates are crucial for accuracy \cite{chen2022cityspec, chen2023cityspec}, but they can add overhead. Enhancing timeliness with text ranking or abstract representation will improve Sim911. (2) \textit{Hallucination}: Despite using advanced LLM backends like GPT-4o, the simulation sometimes generates inaccurate or fabricated information. Further research is needed to mitigate these hallucinations and enhance Sim911's credibility. (3) \textit{Systematic Review on Expert Feedback}: Collecting and systematically reviewing expert feedback during deployment will provide valuable insights for improving Sim911 and other simulation tools. (4) \textit{Model Exploration}: We will explore additional implementations of Sim911 by introducing and fine-tuning more LLMs, such as LlaMa \cite{touvron2023llama} and Falcon \cite{falcon}.


\section{Related Work (Extended)}
\label{sec:extended_related_work}
\textbf{Simulation-based training} is recognized as a pivotal component in preparing individuals for various scenarios across different occupations \cite{suresh2023role, preiksaitis2023opportunities, daun2023systematic}. This training methodology offers a safe and controlled environment for learners to acquire and refine their skills without the real-world consequences of mistakes. Studies show that simulation-based training enhances decision-making, critical thinking, and practical skills, making it an indispensable tool in fields such as healthcare, aviation, and emergency response services, which require high-stakes decision-making \cite{ibrahim2023equity, flores2023artificial, rahman2023emsreact}. Simulation allows for repeated exposure to a wide range of scenarios, some of which may be infrequent or too dangerous to practice in real life, thereby ensuring that trainees are well-prepared for any situation.

In recent years, traditional training setups have begun to evolve, incorporating advanced technologies such as Augmented Reality (AR) and 3D Modeling to provide more \textbf{immersive and interactive learning} experiences \cite{fitria2023augmented, li2018learning, ummenhofer2019lagrangian, pfaff2020learning}. These technological advancements aim to bridge the gap between simulation and real-life experience, offering more realistic and engaging training environments. However, despite these innovations, the core of most training simulations remains largely human-based, relying on scripted scenarios and instructor feedback to guide the learning process \cite{violato2023use, salvato2021crossing, de2020simulation}. This dependence on human-generated content and feedback limits training programs' scalability and adaptability and increases training costs, especially in scenarios that demand rapid updates or customization to address emerging challenges.

\textbf{Large Language Models} (LLMs) demonstrate remarkable performance in generating human-like text, opening new avenues for their application in dialogue-focused simulations \cite{webb2023proof, thoppilan2022lamda, gong2023multimodal}. With their ability to understand and generate natural language, LLMs offer more dynamic and responsive simulation experiences, adapting to the trainee's inputs in real time to create more personalized and unpredictable training scenarios \cite{shanahan2023role}. This capability sets the stage for LLMs to revolutionize training simulations, particularly in domains like emergency call dispatching, where the ability to navigate complex dialogues under pressure is crucial. However, integrating LLMs into training simulations presents challenges, including ensuring the models' responses are accurate, ethical, and free from biases, which are critical considerations for their successful application in important areas \cite{shayegani2023survey, yao2024survey, salewski2024context}.

\textbf{Prompt Engineering} is an important technique in enhancing LLMs' overall performance based on the latest literature \cite{singhal2023towards, wang2023aligning} by using specifically designed workflows based on prompt templates like Chain-of-Thoughts \cite{wei2022chain}, Tree-of-Thoughts \cite{yao2024tree}, Few-shot Prompting \cite{brown2020language, touvron2023llama, kaplan2020scaling}, and Retrieval-Augmented Generation \cite{lewis2020retrieval}. These methodologies leverage structured prompts to guide LLMs in reasoning through complex tasks, enabling more coherent and contextually accurate responses. However, directly applying these techniques in specialized applications such as Sim911, which simulates emergency call handling, results in various performance drops if no relevant knowledge is introduced prior. This phenomenon is well-documented in literature \cite{li2022large} and corroborated by experimental results, indicating that the effectiveness of prompt engineering is highly contingent on the contextual alignment of the prompts with the specific domain knowledge required for the task. Consequently, further refinement and domain-specific adjustments of prompt templates are essential to mitigate performance issues and enhance the utility of LLMs in such specialized scenarios.

\textbf{Text Comparison} metrics are explored and applied in Sim911's evaluation. For our specific goal of measuring realism, authenticity, and equity under emergency response scenarios, traditional metrics like Damerau-Levenshtein distance \cite{damerau1964technique} and BLEU \cite{papineni2002bleu} are unsuitable due to their focus on surface-level text similarities and lack of semantic depth. Other n-gram metrics like ROUGE \cite{lin-2004-rouge} and WER fail to capture the nuances of meaning and context, which are crucial in evaluating emergency response dialogues. While end-to-end metrics such as sentence embeddings and learned metrics \cite{reimers-2019-sentence-bert, cer2018universal, artetxe-etal-2019-laser} do consider semantic content, each of them, when applied in isolation, tends to misalign with our specific criteria for realism and authenticity. Furthermore, these metrics often lack interpretability and generalization in the emergency response context, which is critical for understanding and improving the system's performance. Thus, a comprehensive approach that integrates multiple metrics and adapts them to the emergency response domain is necessary to achieve a more accurate and meaningful evaluation.

\section{Ethical Concerns}
\label{sec:ethic}

Our Institutional Review Board (IRB) has approved the Sim911 project under reference number 232131. We recognize the potential ethical implications of using generative AI technologies, such as large language models (LLMs), in training simulations. Below, we outline how we address these concerns:

\textbf{Human Supervision and Intervention}. Sim911 is a training tool designed exclusively for simulation purposes, aiding 9-1-1 dispatchers without impacting real-world emergency responses. The simulation operates under rigorous human oversight, with training managers actively monitoring all interactions. Both trainers and trainees can terminate the simulation at their discretion.

\textbf{Human Subjects}. Participants in the Sim911 project include 9-1-1 dispatchers and training managers. We refine the system collaboratively through regular feedback sessions, compensating all participants with funds from both institutional and governmental sources. Participants have the freedom to exit the simulation at any point during a session.

\textbf{Privacy and Data Security}. We handle sensitive 9-1-1 call data with utmost care, anonymizing all data to eliminate personal identifiers. Our data storage solutions are secure, utilizing internal cloud servers backed by institutional and government-supported security measures.

\textbf{Bias and Fairness}. We prioritize fairness and actively work to minimize bias, particularly in scenarios involving vulnerable populations. Our approach includes expert consultations in ethics and bias mitigation and ongoing updates to simulation scenarios to ensure diverse and equitable interactions.

\textbf{Continuous Monitoring}. An independent ethics board regularly audits our system, ensuring ethical compliance and optimal performance. These audits help us identify and rectify any issues promptly.

\textbf{Transparency and Accountability}. We maintain a transparent operation, with regular updates and audits shared with stakeholders such as the Chief Information Security Officer (CISO) of the metro government and the Director of the Department of Emergency Communications (DEC). Detailed logs of all simulation activities are recorded and reviewed.

\textbf{Social and Public Safety}. In consultation with key metro government officials, we have implemented additional safeguards, such as tagging data related to vulnerable groups in de-identified call records. Sim911 is strictly a training tool and is not used for any other purposes.

\section{The Design of User Study}
\label{sec:user study design}

We design our user study survey with the following questions and corresponding rating system (not exhaustive list, we only list questions within this paper's scope):
\begin{itemize}
    \item Realism: How similar/vivid are the calls generated by Sim911 to real-world calls? (1-5)
    \item  Authenticity: Are Sim911’s responses valid and true to real-life situation? (1-5)
    \item Inclusiveness: How well does Sim911 provide experiences for simulating vulnerable callers? (1-5)
    \item Helpfulness: How helpful do you think Sim911 assists call-taking training? (1-5)
    \item Improving: Do you observe Sim911's improvements over time? (1-5)
    \item Others: Do you have any comments on realism, authenticity, inclusiveness, helpfulness, and overtime improving? (written)
    \item How well does Sim911 mimic different age groups (e.g., kid, teenager, adult, senior)? (1-5)
    \item How well does Sim911 mimic different emotion ranges (e.g., sad, calm, neutral, anxious, angry, irrational)? (1-5)
    \item How well does Sim911 consider the non-English speaking tag? (1-5)
    \item How well does Sim911 consider the unhoused tag? (1-5)
    \item How well does Sim911 consider the low-income housing area tag? (1-5)
    \item How well does Sim911 consider the mental health issues tag? (1-5)
    \item Others: Do you have any comments on age groups, emotion ranges, non-English speaking, unhoused, low-income housing area, mental health issues want to share? (written)
    \item What do you think of Sim911 compared to traditional human-led training simulations? (less effective, somewhat similar, more effective)
    \item Your training experience is improved by Sim911. (disagree, somewhat, agree)
    \item How satisfied are you with your overall experience with Sim911? (1-5)
\end{itemize}

\section{Running Examples and Algorithms}
\label{sec:algorithms}

\begin{algorithm}[h]
\caption{\small{Context-aware Controlled Generation}}
\label{alg:promptg}
\begin{algorithmic}[1]
\State \textbf{Input:} Constructed Knowledge $\mathcal{D}$, Predefined Tags $\mathcal{T}$, Fact Bases $\mathcal{F}$, Simulation Trials $n$, Pre-configured LLM backends $\mathcal{A}$
\State \textbf{Output:} Generated Prompts $\mathcal{P}$, Specialized Knowledge Bases $\mathcal{K}$, Reference Base $\mathcal{R}$
\For{$i = 0$ \textbf{to} $n$}
    \State Initialize Instructions $\mathcal{I}_{\text{IS}}, \mathcal{I}_{\text{CI}}$ with tags
    \State $\mathcal{T}_{\text{IS}} = \{t_0, \dots, t_k\}$
    \State $\mathcal{T}_{\text{CI}} = \{t_{k+1}, \dots, t_j\}$
    \State Assign $\mathcal{T}_{\text{IS}}$ to $\mathcal{I}_{\text{IS}}$
    \State Assign $\mathcal{T}_{\text{CI}}$ to $\mathcal{I}_{\text{CI}}$
    \ForAll{$t$ \textbf{in} $\mathcal{T}_{\text{IS}}$}
        \ForAll{$d$ \textbf{in} $\mathcal{D}$}
            \If{$d$ \textbf{is associated with} $t$}
                \State Append $d$ to $\mathcal{K}_{\text{IS}}$ \Comment{incident specification}
            \EndIf
        \EndFor
    \EndFor
    \ForAll{$t$ \textbf{in} $\mathcal{T}_{\text{CI}}$}
        \ForAll{$d$ \textbf{in} $\mathcal{D}$}
            \If{$d$ \textbf{is associated with} $t$}
                \State Append $d$ to $\mathcal{K}_{\text{CI}}$ \Comment{caller image construction}
            \EndIf
        \EndFor
    \EndFor
    \State Append \{$\mathcal{I}_{\text{IS}}, \mathcal{F}, \mathcal{K}$\} to References $\mathcal{R}$
\EndFor
\ForAll{$(r, p)$ \textbf{in} $(\mathcal{R}, \mathcal{P})$}
    \State $\mathcal{I}_{\text{IS}}$ selects the most proper LLM agent $a$ \textbf{in} $\mathcal{A}$
    \State $p \gets \text{RAG}(\mathcal{F}, \mathcal{K}_{\text{IS}})$ \Comment{Vector Base Incorporation}
    \State $p \gets \text{CoT}(\mathcal{I}_{\text{IS}})$ \Comment{Detailed Task Explanation}
    \State $p \gets \text{FSP}(\mathcal{K}_{\text{CI}})$ \Comment{Caller Image Deciphering}
    \State $a \gets p$
\EndFor
\end{algorithmic}
\end{algorithm}

\begin{algorithm}[h]
\caption{\small{Validation with Looped Correction}}
\label{alg:validation-loop}
\begin{algorithmic}[1]
\State \textbf{Input:} Generated Response $\mathcal{R}$, LLM Backend $\mathcal{A}$, Loop Threshold $\mathcal{T}$, BERT-based Classifier $\mathcal{B}_{tc}$, RoBERTa-based QA Framework $\mathcal{RB}_{qa}$
\State \textbf{Output:} Validated Response $\mathcal{V}$, User Feedback $\mathcal{F}$

\State $\textit{index} \gets 0$
\State $\mathcal{V} \gets$ \textbf{None}

\While{$\mathcal{R}$ is not validated \textbf{and} $\textit{index} < \mathcal{T}$}
    \State Generate $\mathcal{R}$ using LLM backend $\mathcal{A}$
    \State $\textit{index} \gets \textit{index} + 1$
    
    \If{$\mathcal{R}$ violates format requirements}
        \State Discard $\mathcal{R}$, \textbf{continue} \Comment{\textsc{Format Check}}
    \EndIf

    \State $\hat{\textit{IS}} \gets \mathcal{B}_{tc}(\mathcal{R})$ 
    \If{$\hat{\textit{IS}}$ does not match \textit{IS}}
        \State Discard $\mathcal{R}$, \textbf{continue} \Comment{\textsc{Alignment Check}}
    \EndIf

    \State $\hat{\textit{Address}} \gets \mathcal{RB}_{qa}(\mathcal{R}, \text{`address?'})$ 
    \If{$\hat{\textit{Address}}$ does not exist in real life}
        \State Discard $\mathcal{R}$, \textbf{continue} \Comment{\textsc{Factual Check}}
    \EndIf

    \State $\mathcal{V} \gets \mathcal{R}$
\EndWhile

\If{$\textit{index} = \mathcal{T}$ \textbf{and} $\mathcal{R}$ is not validated} 
    \State Return the best available response
\EndIf

\If{User feedback $\mathcal{F}$ is negative/lower}
    \State Record $\mathcal{F}$ and $\mathcal{V}$; proceed \Comment{\textsc{Human-in-the-loop}}
\EndIf

\end{algorithmic}
\end{algorithm}

\begin{figure*}[h] 
\centering
\includegraphics[width=\textwidth, keepaspectratio=true]{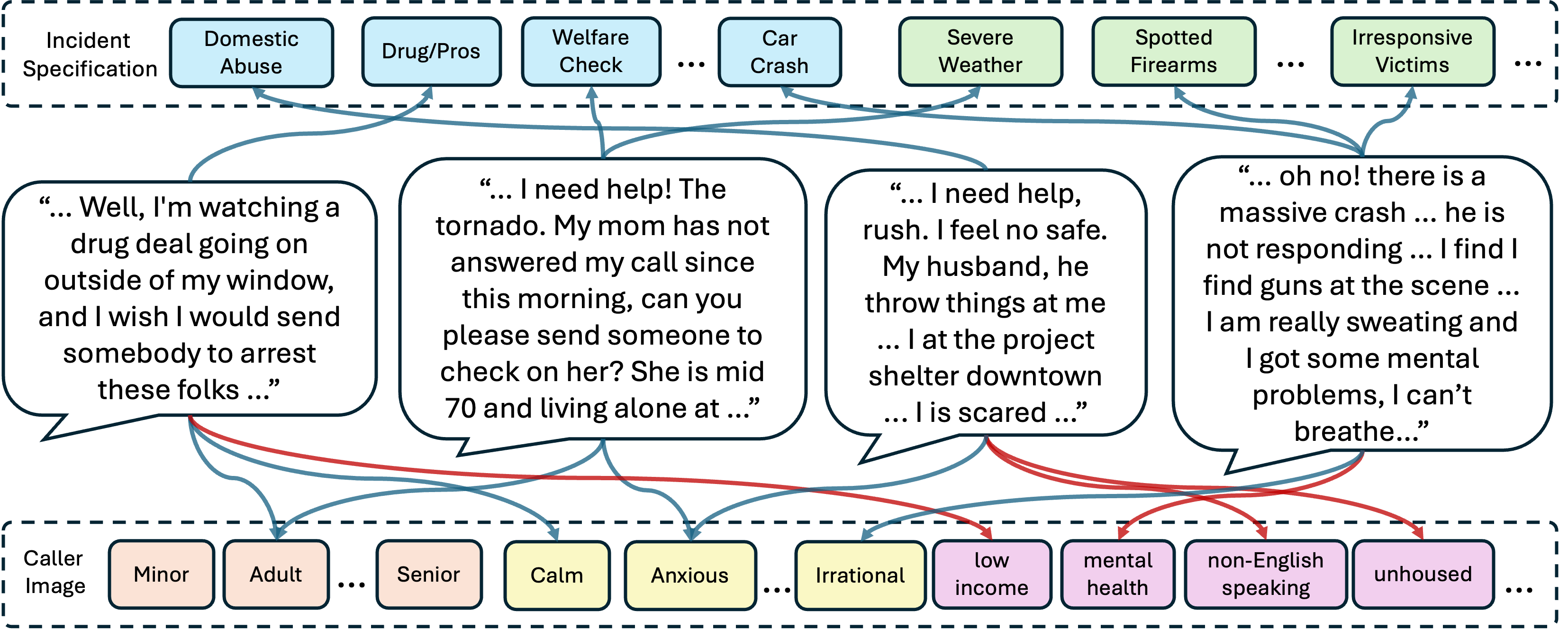}
\caption{\small{Examples on Knowledge Construction process -- introducing tags to existing calls.}} 
\label{fig:knoledge}
\end{figure*}

We provide some running examples to better demonstrate the process of Knowledge Construction, introduced in Section \ref{sec:knowledge}, shown in Figure \ref{fig:knoledge} and Controlled Prompt Generation, introduced in Secion \ref{sec:control}, shown in Figure \ref{fig:prompt}.




As detailed in Algorithm \ref{alg:promptg}, the procedure begins by importing constructed knowledge $\mathcal{D}$, associated tags $\mathcal{T}$, and static fact bases $\mathcal{F}$, setting the stage for a predetermined number of simulations $n$. At the start of each simulation cycle, instruction sets $\mathcal{I}_{\text{IS}}$ and $\mathcal{I}_{\text{CI}}$ are prepared using the relevant tags to accurately depict the incident specifications and caller images. Following this, the system processes the tags to extract and compile necessary data from $\mathcal{D}$ into specialized knowledge bases $\mathcal{K}_{\text{IS}}$ and $\mathcal{K}_{\text{CI}}$, enriching the simulation content and providing a robust contextual backdrop. These compiled datasets and facts are then grouped into reference components $\mathcal{R}$, crucial for crafting the final prompts. Finally, the system selects the most appropriate LLM backend agent $a \in \mathcal{A}$ based on general tags from $\mathcal{I}_{\text{CI}}$. Leveraging CoT, RAG, and FSP, the system dynamically constructs prompts $\mathcal{P}$ and feeds them to $a$. These prompts are designed to not only simulate realistic and varied emergency call situations but also to ensure they align with the tailored needs and profiles identified through the knowledge bases.

\subsubsection{Threshold Adjustment and Backend Selection}
The threshold defines how often Sim911 regenerates responses when failing checks (format, alignment, factual), before being reviewed by human supervisors, as detailed in section Validation with Looped Correction. We adjust the threshold to 3 throughout our deployment considering both latency and negative sample capture rate (the percentage of low-quality responses get filtered out):

\begin{itemize}
    \item T=1: Average latency is 24 ms/token w/ capture rate (CR) of 53.13\%.
    \item T=2 increases latency to 49 ms/token and boosts CR to 77.43\%.
    \item T=3 achieves a 98.47\% CR (similar to T=4 and 5) with shorter latency of 61 ms/token.
\end{itemize}

We also finalized the LLM backend as GPT-4o based on its optimal performance (by the date of the submission): Sim911 with GPT-4o achieved the best realism scores at submission (avg ppl 11.07) compared to LLaMa 3.1 (avg ppl 32.74). Also, many dispatch centers lack resources for local LLM hosting; OpenAI’s API offers an efficient solution.

\section{Archived Call Data \& Training Pieces}
\label{sec:app_calls_training}
In this section, we include a few screenshots of both archived call data and conventional training logs after de-identification. We plan to release the de-identified data samples including both call data and training logs upon the acceptance of this paper.

\subsection{Conventional Training Pieces}

\begin{figure*}[h]
    \centering
    \includegraphics[width=\textwidth, keepaspectratio=true]{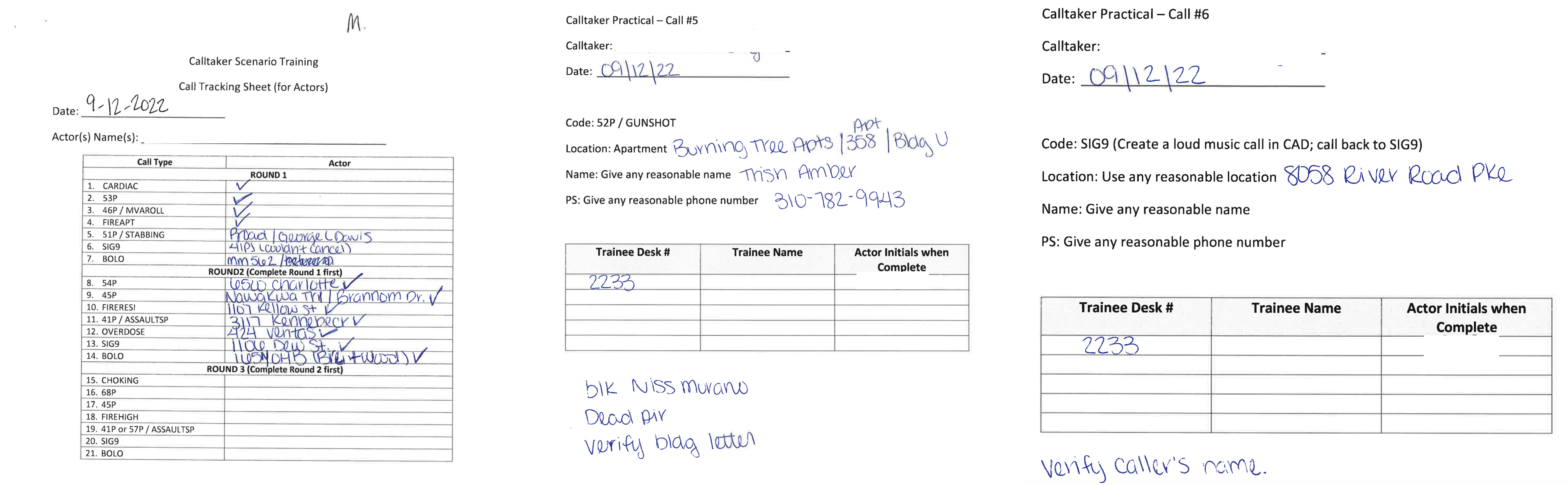}
    \caption{\small{Conventional Training Pieces. }}
    \label{fig:training_pieces}
\end{figure*}

In Figure \ref{fig:training_pieces}, the first page is the training plan, filled by the actor (dispatcher), and the following two parts are the detailed simulation instructions for one specific simulation.

\subsection{Call Data Pieces}

\textbf{[Call Snippet 1]} \\
\textcolor{bettergreen}{911 call-taker}: 9-1-1, what's the address of the emergency?\\
\textcolor{betterblue}{911 caller}: \hl{A 322 Broadway}.\\
\textcolor{bettergreen}{911 call-taker}: Alright, \hl{322 Broadway}, that's downtown.\\
\textcolor{betterblue}{911 caller}: Right in between \hl{3rd} and \hl{4th Avenue South}.\\
\textcolor{bettergreen}{911 call-taker}: What's the phone number you're calling from?\\
\textcolor{betterblue}{911 caller}: \hl{407-456-0938}.\\
\textcolor{bettergreen}{911 call-taker}: What is your name?\\
\textcolor{betterblue}{911 caller}: I'm \hl{Brian Markham-Cammons}.\\
\textcolor{bettergreen}{911 call-taker}: And tell me exactly what happened.\\
\textcolor{betterblue}{911 caller}: I have a homeless fellow named \hl{Jerry} that came in asking us to call an ambulance for him. I'm not sure, what kind of problems are you having \hl{Jerry}? Okay, he said he's hearing voices and then he has pain in his left leg. Hey \hl{Jerry}, how old are you? Fifty-one.\\
\textcolor{bettergreen}{911 call-taker}: Is he breathing normally?\\
\textcolor{betterblue}{911 caller}: Yeah, he's standing up currently, and yeah.\\
\textcolor{bettergreen}{911 call-taker}: Has he been violent at all? Aggressive?\\
\textcolor{betterblue}{911 caller}: No, no, he's not, he's just, he came in asking if we could call him an ambulance.\\
\textcolor{bettergreen}{911 call-taker}: \hl{Jerry}, do you have any weapons on you?\\
\textcolor{betterblue}{911 caller}: No.\\
\textcolor{bettergreen}{911 call-taker}: We have this sent up and they'll be on the way to \hl{322 Broadway} at the \hl{Margaritaville} downtown between \hl{3rd Avenue and 4th Avenue}.\\
\textcolor{bettergreen}{911 call-taker}: Is he going to be waiting inside or outside?\\
\textcolor{betterblue}{911 caller}: He's going to be waiting outside, yeah, I'll be outside.\\
\textcolor{bettergreen}{911 call-taker}: Alright, perfect. I will let them know. If anything changes or gets worse, call us back immediately. In the meantime, just let him rest in the most comfortable position and do not allow him to have anything to eat or drink unless he's diabetic.\\

\textbf{[Call Snippet 2]}\\
\textcolor{bettergreen}{911 call-taker}: Metro \hl{City X} 9-1-1, what is the exact location of your emergency?\\
\textcolor{betterblue}{911 caller}: At \hl{411 Murfreesboro Pike, Apartment 302}.\\
\textcolor{bettergreen}{911 call-taker}: Okay, you said \hl{411 Murfreesboro Pike in South City X, apartment 302}.\\
\textcolor{bettergreen}{911 call-taker}: Okay, is there a building letter or anything with this?\\
\textcolor{bettergreen}{911 call-taker}: Okay, is this near \hl{Paris Avenue} and \hl{Nance Lane} , \hl{37210} is the zip code?\\
\textcolor{bettergreen}{911 call-taker}: Okay, and what's your phone number?\\
\textcolor{betterblue}{911 caller}: \hl{615-946-5518}.\\
\textcolor{bettergreen}{911 call-taker}: And what's your first and last name, ma'am?\\
\textcolor{betterblue}{911 caller}: \hl{Jennifer Robinson}.\\
\textcolor{bettergreen}{911 call-taker}: Okay, and what's going on?\\
\textcolor{betterblue}{911 caller}: I had someone opening my door less than five minutes ago. I had to go to the bathroom and mop my bathroom floor because of some illegal hazardous that had been applied to my walker. And I got it mopped and I came back into the living room and I've got three urine spots that someone has opened the door and placed on it since I had less than three minutes out from the bathroom back to the living room.\\
\textcolor{bettergreen}{911 call-taker}: Okay, do you know who the subject was that did this?\\
\textcolor{betterblue}{911 caller}: No, I do not know, because visually I did not catch them in my apartment, but the door and the lock had to be unlocked, because I keep the door locked at all times. And they're on property and they had to come out of maintenance at \hl{301 or 303}.\\
\textcolor{bettergreen}{911 call-taker}: Okay, but you did not see this subject or anything? But you don't have a description.\\
\textcolor{betterblue}{911 caller}: No, there are small puddles, and they're like puddles, but there were no waters or puddles in the living room. And it's not only one that has access to my front door. It's a family of police officers, and I'm calling it in to you because they're running \hl{illegal and erotic porn} over here, and I have suffered long enough at this property and them not be cracked or turned in by cameras over here.\\
\textcolor{bettergreen}{911 call-taker}: Okay ma'am, the only thing that I can do for you is I can send an officer out there to speak with you about this and you make a report.\\
\textcolor{betterblue}{911 caller}: It's like, oh, well, I'll get it.\\
\textcolor{bettergreen}{911 call-taker}: Yes, I'll get a picture. I can still have it serviced from their body fluids. You did say \hl{411 Murfreesboro Pike in South City X}, \hl{the Mercury Corps Department}, is that correct?\\
\textcolor{betterblue}{911 caller}: That's right, in \hl{apartment 302} for maintenance, \hl{in 302}.\\
\textcolor{bettergreen}{911 call-taker}: Yes ma'am, \hl{302}, I got that. I'll go ahead and send this call up and we'll get an officer out to speak with you, okay?\\
\textcolor{betterblue}{911 caller}: Alright, thank you.\\

\textbf{[Call Snippet 3]}\\
\textcolor{bettergreen}{911 call-taker}: Metro \hl{City X} 9-1-1, what is the location of your emergency?\\
\textcolor{betterblue}{911 caller}: I'll tell you what, I probably ought to call the other one, 865, what is that number? Because we have a suspicious vehicle that's been coming and going right across my business all night, been sitting out by our gate. He just left again, but he only goes for a little while and he comes right back and sits out in front of our gate again. It's a white Nissan, and I don't really, I don't want to go out there because I don't know what they're up to.\\
\textcolor{bettergreen}{911 call-taker}: Okay, what address are you at?\\
\textcolor{betterblue}{911 caller}: \hl{20 Culver Street}, it's \hl{Axel Nobel}.\\
\textcolor{bettergreen}{911 call-taker}: You said \hl{20 Culver Street}? What is his name?\\
\textcolor{betterblue}{911 caller}: \hl{Culvert, C-U-L-V-E-R-T}. I just found out about it, some people have been telling me. He's been there since 10 o'clock last night, and he just, he drives off, but he comes back and sits across from the gate, and he just drove off against us. I'm sure he'll probably come back.\\
\textcolor{bettergreen}{911 call-taker}: What's your first and last name?\\
\textcolor{betterblue}{911 caller}: Uh, \hl{Darren Brown}.\\
\textcolor{bettergreen}{911 call-taker}: And the number you're calling me from?\\
\textcolor{betterblue}{911 caller}: \hl{615-405-4748}.\\
\textcolor{bettergreen}{911 call-taker}: I have a call sent up. We'll get officers in the area. Did you want to speak to officers?\\
\textcolor{betterblue}{911 caller}: Uh, well, I'll tell you what, they can call me and let me know if they have any, if they've seen him or anywhere, but he's riding just right around in this area. So he'll be in this area. Um, you can have one of them to call me if anything goes on, but it's not, don't worry about it.

\section{Snippets of Sim911 Runtimes}

Here we only include three runtimes. We are planning to release all past training interactions as well upon the acceptance of this paper.

\textbf{[Runtime 1]} \\
\textcolor{betterblue}{Sim911}: Hello, I need to report a car that's been parked in front of my house for days. It looks abandoned.\\
\textcolor{bettergreen}{Trainee}: What is your address?\\
\textcolor{betterblue}{Sim911}: My address is \hl{742 Evergreen Terrace}.\\
\textcolor{bettergreen}{Trainee}: Okay, is the car yours?\\
\textcolor{betterblue}{Sim911}: No, the car isn't mine. I've never seen it in the neighborhood before.\\
\textcolor{bettergreen}{Trainee}: How long has the car been there?\\
\textcolor{betterblue}{Sim911}: It's been there for about four days now.\\
\textcolor{bettergreen}{Trainee}: And you say it's a red car. What's the make and model?\\
\textcolor{betterblue}{Sim911}: Actually, I didn't mention the color. The car is a grey sedan, but I'm not sure about the make or model. It's an older model, though.\\
\textcolor{bettergreen}{Trainee}: Is there anyone inside?\\
\textcolor{betterblue}{Sim911}: I haven't seen anyone in the car. It's been empty the whole time.\\
\textcolor{bettergreen}{Trainee}: Are there any other identifying marks, bumper stickers, scars, dents, scratches?\\
\textcolor{betterblue}{Sim911}: Yes, there's a noticeable dent on the passenger side door and a bumper sticker for a local radio station, \hl{WZKB}.\\
\textcolor{bettergreen}{Trainee}: What's the license plate number?\\
\textcolor{betterblue}{Sim911}: I don't have the license plate number. I didn't think to write it down. Should I go check and call you back?\\
\textcolor{bettergreen}{Trainee}: No, that's not necessary. Do you know what state the license plate is from?\\
\textcolor{betterblue}{Sim911}: Yes, the license plate is from our state.\\
\textcolor{bettergreen}{Trainee}: Is there anything suspicious inside the car that you've seen?\\
\textcolor{betterblue}{Sim911}: From what I can see, there doesn't appear to be anything suspicious inside the car. It looks pretty empty.\\
\textcolor{bettergreen}{Trainee}: Okay, what's your name?\\
\textcolor{betterblue}{Sim911}: My name is \hl{Jordan Smith}.\\
\textcolor{bettergreen}{Trainee}: And what's your telephone number, \hl{Jordan}?\\
\textcolor{betterblue}{Sim911}: My phone number is \hl{555-0199}.\\
\textcolor{bettergreen}{Trainee}: Okay, thank you, \hl{Jordan}. Do you want to see police when they come out, or do you just want them to leave?\\
\textcolor{betterblue}{Sim911}: I'm fine with just them checking it out. I don't need to speak with them unless they need more information from me.\\
\textcolor{bettergreen}{Trainee}: Okay, we'll send police. Thank you for calling. Have a good day.

\textbf{[Runtime 2]}\\
\textcolor{betterblue}{Sim911}: Hi, there's a really aggressive driver with red pickup truck here swerving in and out of lanes. It's happening on the \hl{I-40} going east. They're driving extremely dangerously, and I'm really concerned they're going to cause an accident.\\
\textcolor{bettergreen}{Trainee}: Let's see here, \hl{I-40 East} you said, is there a mile marker that you can tell me?\\
\textcolor{betterblue}{Sim911}: Yes, it's right around mile marker 210. This person is driving a red pickup truck, and they're just... they're all over the road.\\
\textcolor{bettergreen}{Trainee}: Trying to find the exact location. Give me one second.\\
\textcolor{bettergreen}{Trainee}: Can I have your number, sir?\\
\textcolor{betterblue}{Sim911}: Yes, my number is \hl{615-555-0123}.\\
\textcolor{bettergreen}{Trainee}: Can I have your full name?\\
\textcolor{betterblue}{Sim911}: My name is \hl{John Smith}.\\
\textcolor{bettergreen}{Trainee}: Alright, \hl{Mr. Smith}, you said it's a red pickup truck, do you know what kind of pickup truck and which direction it's headed?\\
\textcolor{betterblue}{Sim911}: Yes, it's a Ford F-150, and it's headed east on \hl{I-40}.\\
\textcolor{bettergreen}{Trainee}: Where would you like to meet them at?\\
\textcolor{betterblue}{Sim911}: Could we meet at the parking lot of the \hl{City X Public Library} on \hl{Church Street}? It's a public space, and I'll be able to get there in about 15 minutes.\\
\textcolor{bettergreen}{Trainee}: Okay, alright \hl{Mr. Smith}, I got a call set up and you said that was mile marker 210 on \hl{I-40 East} and I will get this set up and the officer will meet you at the \hl{City X Public Library} and I've got everything I need. Is there anything else you need?\\
\textcolor{betterblue}{Sim911}: No, that should be everything. Thank you so much for your help. I really appreciate it.\\
\textcolor{bettergreen}{Trainee}: Okay, got it. Goodbye.

\textbf{[Runtime 3]}\\
\textcolor{betterblue}{Sim911}: Hi, I need to report some property damage that happened at my place. Can you help me with this?\\
\textcolor{bettergreen}{Trainee}: Yeah, can you go ahead and give me the address?\\
\textcolor{betterblue}{Sim911}: It's \hl{742 Evergreen Terrace, in City X}.\\
\textcolor{bettergreen}{Trainee}: That's not showing up as an address for me. Can you give me another address?\\
\textcolor{betterblue}{Sim911}: Sorry for the confusion, it's \hl{Evergreen Terrace}.\\
\textcolor{bettergreen}{Trainee}: Good.\\
\textcolor{betterblue}{Sim911}: Thank you for understanding. So, there was some damage to my front lawn and mailbox. It looks like someone drove through it last night.\\
\textcolor{betterblue}{Sim911}: Hello, are you still there? Do you need any more information to help with this situation?\\
\textcolor{bettergreen}{Trainee}: Can I get your first and last name?\\
\textcolor{betterblue}{Sim911}: Of course, my name is \hl{John Smith}.\\
\textcolor{bettergreen}{Trainee}: Is there, do you see tire marks?\\
\textcolor{betterblue}{Sim911}: Yes, there are very clear tire marks across the lawn and leading up to the damaged mailbox.\\
\textcolor{bettergreen}{Trainee}: Alright, and then would you like to speak to officers when they arrive?\\
\textcolor{betterblue}{Sim911}: Yes, I would appreciate speaking to the officers when they arrive to provide any additional details they might need.\\
\textcolor{bettergreen}{Trainee}: All right.\\
\textcolor{betterblue}{Sim911}: Thank you for your help. Is there anything else you need from me at this stage?\\
\textcolor{bettergreen}{Trainee}: Is there anyone that's injured that you can see?\\
\textcolor{betterblue}{Sim911}: No, there are no injuries. It looks like it was just property damage.\\
\textcolor{bettergreen}{Trainee}: All right, we will have that place checked soon. Bye.\\

\section{Co-pilot Designs in In-Context Validation}
\label{sec:copilot}

\textbf{BERT Classifier for Incident Type}
Given a sequence of conversational turns $\{T_1, T_2, \ldots, T_n\}$, the BERT classifier aims to predict the type of incident $C$ associated with the conversation. Each turn $T_i$ is represented as a tokenized input to BERT, which processes the input to predict the incident class. The classification process can be formulated as:

\begin{equation}
    C = \text{BERT}_{\text{class}}(T_1, T_2, \ldots, T_n)
\end{equation}

where $\text{BERT}_{\text{class}}$ represents the BERT model fine-tuned for classification tasks.

\textbf{RoBERTa-based Question Answering}
The RoBERTa model is utilized to determine whether a specific turn in the conversation contains an answer to a given question $Q$. For each turn $T_i$, the model evaluates the presence of an answer by processing both the turn and the question together. The output is a binary indicator $A_i$, where $A_i = 1$ signifies that the turn contains an answer, and $A_i = 0$ otherwise. This can be represented as:

\begin{equation}
    A_i = \text{RoBERTa}_{\text{QA}}(T_i, Q)
\end{equation}

where $\text{RoBERTa}_{\text{QA}}$ denotes the RoBERTa model adapted for the question-answering task.

\section{Background of Evaluation Metrics}
\label{sec: eval_bg}

In this section, we introduce each of the mentioned metrics in evaluation in detail.

\subsection{METEOR}

METEOR (Metric for Evaluation of Translation with Explicit ORdering) is a metric used to evaluate the quality of machine-generated text, especially in the context of machine translation and natural language generation. It aims to address some of the limitations of the BLEU metric by incorporating synonymy, stemming, and paraphrasing, and by emphasizing precision, recall, and a harmonic mean.

\textbf{Definition of METEOR}. Unlike BLEU, which primarily focuses on precision, METEOR considers both precision and recall, thereby providing a more balanced evaluation of translation quality. METEOR computes an alignment between the candidate and reference translations based on exact matches, stemmed matches, synonym matches, and paraphrase matches.

\textbf{Calculation of METEOR}. The METEOR score is calculated through the following steps:
\begin{enumerate}
    \item \textbf{Tokenization:} Both the candidate translation and reference translations are tokenized into words.
    \item \textbf{Matching:} Tokens from the candidate translation are matched to tokens in the reference translation using exact matches, stem matches, synonym matches, and paraphrase matches.
    \item \textbf{Precision and Recall:} The precision ($P$) and recall ($R$) are calculated based on the matches.
\end{enumerate}

\begin{align}
    P = \frac{m}{|c|}\\
    R = \frac{m}{|r|}
\end{align}

where:
\begin{itemize}
    \item $m$ is the number of matched tokens.
    \item $|c|$ is the total number of tokens in the candidate translation.
    \item $|r|$ is the total number of tokens in the reference translation.
\end{itemize}

\textbf{Harmonic Mean and Penalty}. The harmonic mean of precision and recall, also known as the F-score, is computed as follows:

\begin{align}
    F_{mean} = \frac{10 \cdot P \cdot R}{R + 9 \cdot P}
\end{align}

To account for the fragmentation of matches, METEOR introduces a penalty ($P_{\text{frag}}$):

\begin{align}
P_{\text{frag}} = 0.5 \left( \frac{num\_chunks}{m} \right) 
\end{align}

where $num\_chunks$ is the number of contiguous sequences of matched tokens. The final METEOR score is calculated by combining the harmonic mean and the penalty:

\begin{align}
    \text{METEOR} = F_{mean} \cdot (1 - P_{\text{frag}})
\end{align}

\textbf{Interpretation of METEOR Scores}. METEOR scores range from 0 to 1, where higher scores indicate better alignment between the candidate and reference translations. The metric provides a more nuanced evaluation than BLEU by considering recall, synonyms, and paraphrases, making it particularly useful for assessing translation quality in a more human-like manner.

\subsection{Perplexity (PPL)}

Perplexity is a metric commonly used to evaluate the performance of language models in natural language processing (NLP). It measures how well a probabilistic model predicts a sample and is often used to compare different language models.

\textbf{Definition of Perplexity}. Perplexity is defined as the exponentiated average negative log-likelihood of a sequence of words. For a language model, it is a measure of how surprised the model is by the test data, with lower perplexity indicating a better model. Formally, for a given sequence of words $w_1, w_2, \ldots, w_N$, the perplexity ($PP$) is given by:

\begin{align}
    PP(w_1, w_2, \ldots, w_N) 
    &= \exp\left( 
    -\frac{1}{N} 
    \sum_{i=1}^{N} 
    \log P(w_i \mid \right. \nonumber \\
    &\quad \left. w_1, w_2, \ldots, w_{i-1})
    \right)
\end{align}

\textbf{Calculation of Perplexity}. The calculation involves the following steps:

\begin{enumerate}
    \item \textbf{Log-Likelihood:} Calculate the log-likelihood of each word in the sequence given the previous words.
    \item \textbf{Average Log-Likelihood:} Compute the average of these log-likelihoods over the entire sequence.
    \item \textbf{Exponentiation:} Exponentiate the negative average log-likelihood to obtain the perplexity.
\end{enumerate}

Given a language model that assigns a probability $P(w_i \mid w_1, w_2, \ldots, w_{i-1})$ to each word $w_i$ in the sequence, the perplexity is calculated as:

\begin{align}
    PP &= \exp\left(-\frac{1}{N} \sum_{i=1}^{N} \log P(w_i \mid w_1, w_2, \ldots, \right. \nonumber \\
    &\quad \left. w_{i-1})\right)
\end{align}

Alternatively, if the cross-entropy $H$ of the model on the sequence is known, the perplexity can be computed as:

\begin{align}
    PP = \exp(H)
\end{align}

where the cross-entropy $H$ is given by:

\begin{align}
    H = -\frac{1}{N} \sum_{i=1}^{N} \log P(w_i \mid w_1, w_2, \ldots, w_{i-1})
\end{align}

\textbf{Interpreting Perplexity Scores}. Perplexity scores provide an intuitive measure of a language model's performance:
\begin{itemize}
    \item \textbf{Lower Perplexity:} Indicates that the model is better at predicting the test data, implying higher quality and accuracy.
    \item \textbf{Higher Perplexity:} Suggests that the model struggles to predict the test data, indicating lower quality.
\end{itemize}

For example, if a language model assigns high probabilities to the actual words in the sequence, it will have a lower perplexity, reflecting better performance. Conversely, if the model assigns low probabilities, the perplexity will be higher, indicating poorer performance.

\subsection{Typical-Token Ratio (TTR)}

Type-Token Ratio (TTR) is a measure used in linguistics and natural language processing (NLP) to assess the lexical diversity of a text. It is calculated by dividing the number of unique words (types) by the total number of words (tokens) in a given text. A higher TTR indicates greater lexical diversity.

\textbf{Definition of Type-Token Ratio}. Formally, let $V$ be the number of unique words (types) and $N$ be the total number of words (tokens) in a text. The Type-Token Ratio (TTR) is defined as:

\begin{align}
    \text{TTR} = \frac{V}{N}
\end{align}

\textbf{Calculation of Type-Token Ratio}. The calculation involves the following steps:

\begin{enumerate}
    \item \textbf{Count Tokens:} Count the total number of words (tokens) in the text.
    \item \textbf{Count Types:} Count the number of unique words (types) in the text.
    \item \textbf{Compute TTR:} Divide the number of types by the number of tokens.
\end{enumerate}

Given a text sample, the TTR can be computed as follows:

\begin{align}
    \text{TTR} = \frac{\text{Number of Unique Words (Types)}}{\text{Total Number of Words (Tokens)}}
\end{align}

\textbf{Interpreting TTR Scores}. TTR scores provide insight into the lexical diversity of a text:
\begin{itemize}
    \item \textbf{High TTR:} A high TTR indicates a text with a large variety of unique words relative to its length, suggesting greater lexical diversity.
    \item \textbf{Low TTR:} A low TTR suggests that the text has many repeated words relative to its length, indicating lower lexical diversity.
\end{itemize}

\textbf{Considerations}. While TTR is a useful measure, it can be sensitive to the length of the text. Longer texts tend to have lower TTRs because the likelihood of repeating words increases.

\subsection{Map Matching}

This section outlines the workflow for processing location information obtained from conversations, including the steps for manual review and validation using different backend services such as Google Maps and OpenStreetMap. The workflow ensures accurate location identification and validation, particularly when an absolute match is required.

\textbf{Workflow Steps}. The workflow consists of the following steps:

\begin{enumerate}
    \item \textbf{Obtain the Utterance with Location Information:} Extract the utterance containing location information from the conversation.
    \item \textbf{Manual Review and Formatting:} Manually review the extracted utterance to obtain a well-formatted address, including block numbers and street names if possible.
    \item \textbf{Backend Processing with Google Maps:} 
    \begin{enumerate}
        \item Pass the raw utterance directly to Google Maps.
        \item Check if Google Maps returns a valid GPS location.
    \end{enumerate}
\end{enumerate}

\textbf{Detailed Workflow Description}. Step 1: Obtain the Utterance with Location Information: The first step involves extracting the utterance from the conversation that contains the location information. Step 2: Manual Review and Formatting. Once the location information is extracted, it is manually reviewed to ensure accuracy. The goal is to obtain a well-formatted address that includes block numbers and street names if possible. This formatted address will be used in subsequent steps. Step 3: Backend Processing with Google Maps: If the backend service is Google Maps, the raw utterance is directly passed to Google Maps API. The response is checked to determine if a valid GPS location is returned..

\subsection{Sytax Overlap Analysis using Context-Free Grammar Parser}

Context-Free Grammar (CFG) parsers are fundamental tools in computational linguistics and natural language processing (NLP). They are used to analyze the syntactic structure of sentences based on a set of production rules. When applied to syntax overlap analysis, CFG parsers can help identify and compare the syntactic patterns between different texts, providing insights into their structural similarities and differences.

\textbf{Context-Free Grammar (CFG) Overview}. A Context-Free Grammar (CFG) is a formal grammar that consists of a set of production rules used to generate all possible strings in a given formal language. A CFG is defined by a 4-tuple $(N, \Sigma, P, S)$, where:
\begin{itemize}
    \item $N$ is a finite set of non-terminal symbols.
    \item $\Sigma$ is a finite set of terminal symbols.
    \item $P$ is a finite set of production rules of the form $A \rightarrow \beta$, where $A \in N$ and $\beta \in (N \cup \Sigma)^*$.
    \item $S$ is the start symbol.
\end{itemize}

To apply a CFG parser in syntax overlap analysis, follow these steps:

\begin{enumerate}
    \item \textbf{Parse the Sentences:} Use the CFG parser to generate parse trees for each sentence. The parse tree represents the syntactic structure of the sentence according to the grammar rules.
    \item \textbf{Extract Syntactic Structures:} Extract the syntactic structures from the parse trees, focusing on key elements such as noun phrases (NP), verb phrases (VP), and other grammatical constituents.
    \item \textbf{Compare Structures:} Compare the syntactic structures of the sentences to identify overlaps and differences. This can be done by matching corresponding subtrees or syntactic patterns between the parse trees.
\end{enumerate}

\textbf{Detailed Steps}.

\textit{Step 1: Parse the Sentences}. Given two sentences $S_1$ and $S_2$, use a CFG parser to generate their respective parse trees $T_1$ and $T_2$.

\begin{align}
    T_1 = \text{CFG\_Parser}(S_1)\\
    T_2 = \text{CFG\_Parser}(S_2)
\end{align}

\textit{Step 2: Extract Syntactic Structures}. From the parse trees $T_1$ and $T_2$, extract the syntactic structures such as noun phrases (NP), verb phrases (VP), and other constituents.

\textit{Step 3: Compare Structures}. Compare the extracted syntactic structures to measure the overlap. For instance, calculate the similarity between corresponding subtrees using tree edit distance or other similarity measures.

\subsection{Lexicon Overlap Analysis using TF-iDF}

Lexicon overlap analysis is a method used to evaluate the similarity between texts based on their lexical content. One effective way to perform this analysis is by using Term Frequency-Inverse Document Frequency (TF-iDF), a statistical measure that reflects the importance of a word in a document relative to a corpus. TF-iDF helps in identifying and comparing the significant terms between texts, providing insights into their lexical overlap and distinctiveness.

\textbf{Term Frequency-Inverse Document Frequency (TF-iDF) Overview}. TF-iDF is a numerical statistic that is intended to reflect how important a word is to a document in a collection or corpus. It is composed of two components: Term Frequency (TF) and Inverse Document Frequency (iDF).

\textbf{Term Frequency (TF)} Term Frequency measures how frequently a term appears in a document. It is calculated as:

\begin{align}
    \text{TF}(t, d) = \frac{f_{t,d}}{N_d}
\end{align}

where:
\begin{itemize}
    \item $f_{t,d}$ is the frequency of term $t$ in document $d$.
    \item $N_d$ is the total number of terms in document $d$.
\end{itemize}

\textbf{Inverse Document Frequency (iDF)}. Inverse Document Frequency measures the importance of a term by reducing the weight of terms that appear frequently across all documents in the corpus. It is calculated as:

\begin{align}
    \text{iDF}(t, D) = \log \left( \frac{N}{|d \in D : t \in d|} \right)
\end{align}

where:
\begin{itemize}
    \item $N$ is the total number of documents in the corpus $D$.
    \item $|d \in D : t \in d|$ is the number of documents containing term $t$.
\end{itemize}

\textbf{TF-iDF Calculation}. The TF-iDF score for a term $t$ in a document $d$ is the product of its TF and iDF values:

\begin{align}
    \text{TF-iDF}(t, d, D) = \text{TF}(t, d) \times \text{iDF}(t, D)
\end{align}

\textbf{Lexicon Overlap Analysis Using TF-iDF}. To perform lexicon overlap analysis using TF-iDF, follow these steps:

\begin{enumerate}
    \item \textbf{Compute TF-iDF Scores:} Calculate the TF-iDF scores for all terms in the documents.
    \item \textbf{Identify Significant Terms:} Identify the significant terms in each document based on their TF-iDF scores.
    \item \textbf{Compare Lexical Content:} Compare the sets of significant terms between documents to measure lexical overlap.
\end{enumerate}

\textit{Step 1: Compute TF-iDF Scores}

Given a set of documents $D = \{d_1, d_2, \ldots, d_n\}$, compute the TF-iDF score for each term $t$ in each document $d_i$.

\textit{Step 2: Identify Significant Terms}

For each document $d_i$, select the terms with the highest TF-iDF scores as the significant terms. These terms are likely to be the most important words in the document.

\textit{Step 3: Compare Lexical Content}

Compare the sets of significant terms between documents to measure the lexical overlap. This can be done using various similarity measures such as Jaccard similarity:

\begin{align}
   \text{Jaccard Similarity}(A, B) = \frac{|A \cap B|}{|A \cup B|} 
\end{align}

where $A$ and $B$ are the sets of significant terms from two documents.

\subsection{Sentiment Overlap Analysis using TextBlob}

Sentiment overlap analysis is a method used to compare the sentiment expressed in different texts. This analysis is important in various natural language processing (NLP) applications, such as opinion mining, customer feedback analysis, and social media monitoring. By examining the sentiment overlap, one can assess the consistency and similarity of sentiments expressed across multiple texts. TextBlob is a popular Python library for processing textual data that includes tools for sentiment analysis, making it an excellent choice for this task.

\textbf{TextBlob for Sentiment Analysis}. TextBlob is a Python library for processing textual data. It provides a simple API for common NLP tasks such as part-of-speech tagging, noun phrase extraction, translation, and importantly, sentiment analysis. TextBlob's sentiment analysis tool provides polarity and subjectivity scores for a given text.

\textbf{Polarity and Subjectivity}. TextBlob provides two main sentiment metrics:
\begin{itemize}
    \item \textbf{Polarity:} A float value within the range [-1.0, 1.0], where -1.0 indicates a very negative sentiment, 1.0 indicates a very positive sentiment, and 0 indicates a neutral sentiment.
    \item \textbf{Subjectivity:} A float value within the range [0.0, 1.0], where 0.0 is very objective and 1.0 is very subjective.
\end{itemize}

\textbf{Methodology for Sentiment Overlap Analysis}. To perform sentiment overlap analysis using TextBlob, follow these steps:

\begin{enumerate}
    \item \textbf{Sentiment Extraction:} Use TextBlob to extract the sentiment scores (polarity and subjectivity) for each text.
    \item \textbf{Comparison:} Compare the sentiment scores to measure the overlap and similarity between the texts.
\end{enumerate}

\textit{Step 1: Sentiment Extraction} For each text, use TextBlob to obtain the sentiment scores. Given a set of texts $T = \{t_1, t_2, \ldots, t_n\}$, the sentiment scores for each text $t_i$ are computed as:

\begin{align}
   \text{Sentiment}(t_i) = (\text{Polarity}(t_i), \text{Subjectivity}(t_i)) 
\end{align}

\textit{Step 2: Comparison}. To compare the sentiment scores, calculate the overlap and similarity between the sentiment vectors of the texts. One approach is to use cosine similarity to measure the similarity between the sentiment vectors:

\begin{align}
    \text{Cosine Similarity}(\mathbf{S}_1, \mathbf{S}_2) = \frac{\mathbf{S}_1 \cdot \mathbf{S}_2}{\|\mathbf{S}_1\| \|\mathbf{S}_2\|}
\end{align}

where $\mathbf{S}_1$ and $\mathbf{S}_2$ are the sentiment vectors of two texts.